\documentclass[dvipsnames]{article}
\usepackage{template_conference}
\usepackage[utf8]{inputenc} 
\usepackage[T1]{fontenc}    
\usepackage{hyperref}       
\usepackage{url}            
\usepackage{booktabs}       
\usepackage{amsfonts}       
\usepackage{amsmath}        
\usepackage{amssymb}        
\usepackage{amsthm}         
\usepackage{nicefrac}       
\usepackage{microtype}      
\usepackage{xcolor}         
\usepackage{graphicx}       
\usepackage{array}          
\usepackage{multirow}       
\usepackage{enumitem}       
\usepackage{wrapfig}        
\usepackage{pifont}         
\usepackage{listings}       

\usepackage{marvosym}

\definecolor{promptbg}{RGB}{248,248,245}
\definecolor{promptframe}{RGB}{200,200,200}
\definecolor{promptkw}{RGB}{30,90,170}
\definecolor{promptstr}{RGB}{150,60,30}
\definecolor{promptcmt}{RGB}{110,110,110}
\lstdefinestyle{promptstyle}{
  language=Python,
  basicstyle=\ttfamily\scriptsize,
  keywordstyle=\color{promptkw}\bfseries,
  stringstyle=\color{promptstr},
  commentstyle=\color{promptcmt}\itshape,
  backgroundcolor=\color{promptbg},
  frame=single,
  rulecolor=\color{promptframe},
  framesep=4pt,
  breaklines=true,
  breakatwhitespace=false,
  breakindent=0pt,
  postbreak=\mbox{\textcolor{promptframe}{$\hookrightarrow$}\space},
  columns=fullflexible,
  keepspaces=true,
  showstringspaces=false,
  upquote=true,
  literate={“}{{``}}1 {”}{{''}}1 {‘}{{`}}1 {’}{{'}}1,
  aboveskip=4pt,
  belowskip=4pt,
}

\definecolor{gaingreen}{RGB}{30,120,60}
\newcommand{\subgain}[1]{%
  \ensuremath{_{\textcolor{gaingreen}{\scriptstyle #1}}}%
}

\newtheorem{definition}{Definition}
\newtheorem{property}{Property}
\newtheorem{proposition}{Proposition}

\setcitestyle{numbers,square,comma}
\usepackage{tcolorbox}
\newtcolorbox{abstractbox}{
  colback=blue!5!white, frame empty, boxrule=1pt, arc=4mm,
  left=8pt, right=8pt, top=8pt, bottom=8pt, opacityback=0.9
}
\renewenvironment{abstract}{\begin{abstractbox}\begin{center}\textbf{\Large Abstract}\end{center}}{\end{abstractbox}}
\title{Med-Agentic: Distilling Agentic Medical Reasoning\\with Internalized Meta-Capabilities}
\author{
\footnotesize
\textbf{Yucheng Zhou}\textsuperscript{*}\quad
\textbf{Junwei Sheng}\textsuperscript{*}\quad
\textbf{Jianbing Shen}\textsuperscript{\Letter}\\[3pt]
\scriptsize SKL-IOTSC, Department of AIX, University of Macau\\[2pt]
\texttt{yucheng.zhou@connect.um.edu.mo}
}
\begin{document}
\maketitle
\begingroup
\renewcommand{\thefootnote}{\fnsymbol{footnote}}
\footnotetext[1]{Equal contribution.}
\endgroup
\begingroup
\renewcommand{\thefootnote}{\Letter}
\footnotetext{Corresponding author: \textit{Jianbing Shen}.}
\endgroup
\begin{abstract}
Large language models (LLMs) in expert domains face a trade-off between \emph{agentic tool use}, which externalizes expertise behind brittle multi-turn runtimes, and \emph{end-to-end specialization}, which internalizes expertise only implicitly at task-level granularity. We propose \emph{internalized meta-capabilities}: atomic, parametrically internalized, and composable reasoning units that an LLM can select and combine within a single chain-of-thought (CoT). We instantiate this idea as \textbf{Med-Agentic} for medical visual diagnosis, where the model emits a capability-tagged structured CoT in one forward pass, turning capability invocation into tagged token generation rather than external tool calls. For training, \textbf{Med-OPD} performs multi-teacher on-policy distillation with \emph{token-level dual-factor teacher routing} conditioned on both the image and the active capability tag, strictly generalizing per-prompt teacher routing. We show that this routing gives an unbiased estimator of the same expected teacher-KL gradient. Across public medical imaging datasets spanning nine anatomies and eight modalities, Med-Agentic consistently improves in-distribution and out-of-distribution diagnostic accuracy over the strong competitors.
\end{abstract}

\section{Introduction}
\label{sec:intro}

Large language models, including their multimodal variants (MLLMs), are
increasingly deployed in expert domains such as medicine, law, engineering, and scientific analysis. In these settings, the dominant
practice oscillates between two paradigms, neither of which is fully
satisfactory. The first treats specialized capabilities as
\emph{external} tools and lets the LLM act as a planner that
orchestrates them through multi-turn interactions, as exemplified by
ReAct~\citep{DBLP:conf/iclr/YaoZYDSN023}, Toolformer~\citep{DBLP:conf/nips/SchickDDRLHZCS23}, and the medical pipeline
MedAgent-Pro~\citep{DBLP:journals/corr/abs-2503-18968}. Such systems are compositional in spirit, but
their expertise lives outside the model: it is brittle to API or
environment shifts, incurs latency through repeated external calls, and
is not improved by training. The second paradigm collapses the entire
input-output mapping into a single fine-tuned model, either at the
level of individual diseases or via large-scale multi-task supervised
fine-tuning~\citep{DBLP:conf/iclr/WeiBZGYLDDL22}. The resulting models are fast and self-contained,
but their internal expertise is implicit and tied to the labels seen
during training: they tend to learn ``input $\to$ conclusion'' rather
than ``how to observe'', and generalization to unseen diseases or
modalities typically requires retraining.

Both paradigms overlook a basic fact about expert reasoning: clinicians
do not memorize a separate routine for each disease. Instead, they
master a small repertoire of \emph{observational primitives}, color,
density, morphology, geometry, margin, spatial relationship, and so on,
and \emph{compose} them when confronting a new case~\citep{DBLP:journals/corr/abs-2505-24173}. The same
primitive ``irregular margin'' is used to describe a lung nodule, a
skin lesion, and a breast mass; the diagnostic process is the dynamic
selection and composition of a small set of reusable observations,
not the recall of an indivisible disease-specific procedure. This is
precisely the kind of structure that compositional generalization
theory~\citep{DBLP:journals/corr/abs-2302-01067} predicts will transfer to novel combinations.

\begin{figure}[t]
  \centering
  \includegraphics[trim=0 0 95bp 0,clip,width=\linewidth]{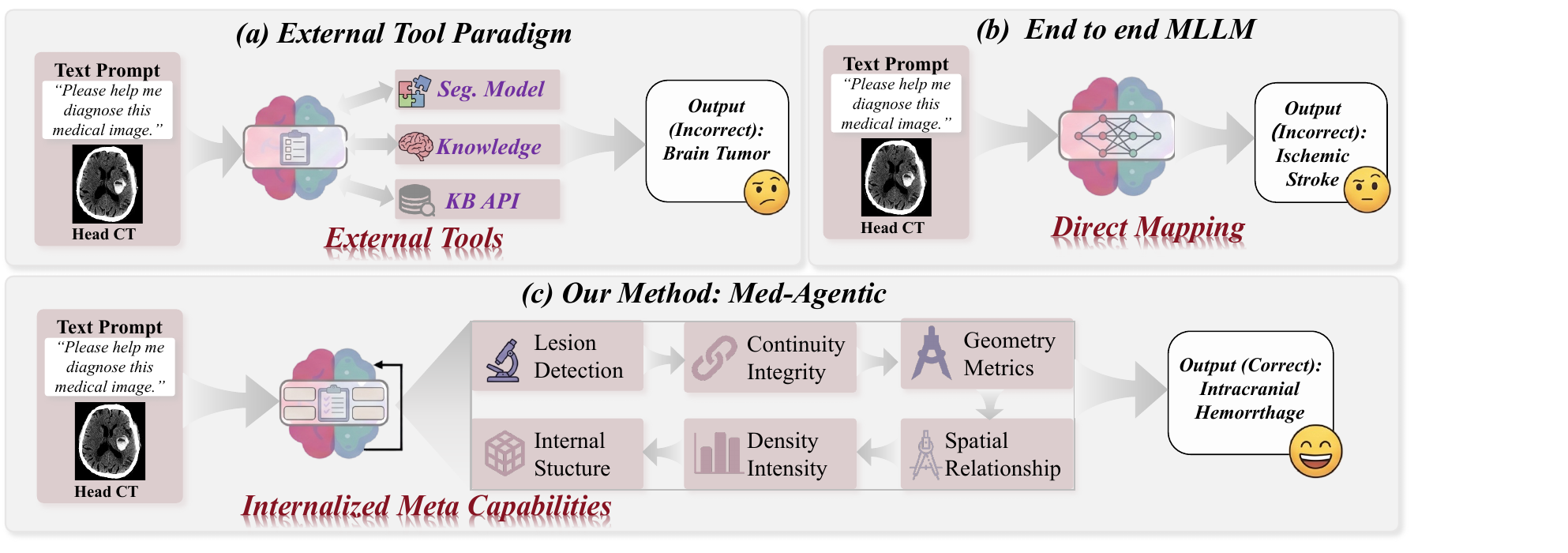}
  \caption{\small Three paradigms for deploying LLMs in expert domains. Med-Agentic
  realizes agentic capability composition as a single-pass capability-tagged
  CoT, without external tool runtime or multi-turn scaffolding. Capabilities
  are internalized in the LLM parameters and selected dynamically at the
  token level.}
  \label{fig:teaser}
\end{figure}

Motivated by this observation, we argue that an LLM in an expert domain
should acquire a set of \emph{internalized meta-capabilities}: atomic
reasoning units that are simultaneously (i) \emph{atomic} (each
captures one indivisible observational dimension and is reused across
tasks), (ii) \emph{internalized} (implemented by the model's own
parameters, so that invocation is a generation step rather than an
external API call), and (iii) \emph{composable} (the LLM agentically
selects and combines them within a single chain of thought, and the
composition policy itself is learnable). Internalized meta-capabilities
are neither tools nor tasks but a new abstraction layer between the
two, made precise in Section~\ref{sec:framework}. We instantiate this
paradigm on medical visual diagnosis as \textbf{Med-Agentic}
(Figure~\ref{fig:teaser}). Concretely, we (a) decompose visual
diagnosis into $K{=}14$ clinical observational meta-capabilities; (b)
expose agentic capability invocation through a single-pass interface,
\emph{capability-tagged structured CoT}, in which each
$\langle\textsc{think}\rangle$ segment is prefixed by a capability key
drawn from the existing vocabulary, so that ``calling a capability''
becomes a token sequence rather than a multi-turn protocol; and (c)
train the student in three stages, namely per-capability SFT,
structured-CoT SFT, and multi-teacher on-policy distillation
(\textbf{Med-OPD}), which forges four modality/anatomy specialists from
a shared cold-start checkpoint and distills them into one student via
\emph{token-level dual-factor} teacher routing that conditions jointly
on the image embedding and the active capability tag.

We evaluate Med-Agentic on $30$ public medical imaging datasets
covering $9$ anatomies and $8$ modalities ($119{,}373$ samples), with a
$6{,}033$-sample in-distribution test set and three held-out
out-of-distribution slices that probe cross-modality, label-shift, and
emergent-disease transfer. Under a shared Qwen3.5-9B~\citep{DBLP:journals/corr/abs-2502-13923} backbone
and identical training pool, Med-Agentic improves top-$1$ accuracy
over the strongest trained baseline by $+3.9$ points on
in-distribution test and by $+6.2$ points averaged over the three OOD
slices, while widening its lead on samples that require composing
four or more capabilities. The contributions of this work are as
follows.

\begin{itemize}[leftmargin=1.4em,itemsep=2pt,topsep=2pt]
\item We introduce \emph{internalized meta-capabilities} as a third paradigm
between agentic tool use and end-to-end specialization, and
formalize the three properties (atomicity, internalization,
compositionality) that distinguish it from both prior paradigms.
\item We propose \emph{capability-tagged structured CoT}, a single-pass
interface that compiles agentic capability invocation into a token
sequence over the model's existing vocabulary and admits segment-level
supervision without any external runtime.
\item We design \textbf{Med-OPD}, a multi-teacher on-policy
distillation procedure for multimodal, structured-CoT trajectories
whose central methodological novelty is a \emph{token-level dual-factor}
teacher routing that strictly generalizes per-prompt static routing
while remaining an unbiased estimator of the same expected teacher KL
gradient.
\item We instantiate the paradigm as Med-Agentic, the first medical
MLLM to internalize and agentically compose a clinical capability
taxonomy in a single forward pass, and demonstrate consistent gains
over end-to-end, agentic-tool, and on-policy RL baselines on both
in-distribution and out-of-distribution diagnostic transfer.
\end{itemize}

\section{Related Work (summary)}
\label{sec:related}
Existing agentic systems such as ReAct~\citep{DBLP:conf/iclr/YaoZYDSN023}, Toolformer~\citep{DBLP:conf/nips/SchickDDRLHZCS23}, ARTIST~\citep{DBLP:journals/corr/abs-2505-01441}, ReTool~\citep{DBLP:journals/corr/abs-2504-11536}, role-specialized multi-agent frameworks such as MAGIS~\citep{DBLP:conf/nips/0003ZWZ0C24}, and medical variants including Octopus~\citep{DBLP:journals/corr/abs-2511-15351} and MedAgent-Pro~\citep{DBLP:journals/corr/abs-2503-18968} externalize expertise as LLM-planned tool orchestration, whereas Med-Agentic internalizes reusable clinical capabilities as tagged CoT segments. Related internalization work such as OneThinker~\citep{DBLP:journals/corr/abs-2512-03043} and Skill0~\citep{lu2026skill0} supports specialized abilities but remains coarse-grained or lacks an atomic, compositional, multimodal capability interface. For optimization, RLHF~\citep{DBLP:conf/nips/Ouyang0JAWMZASR22} and GRPO~\citep{DBLP:journals/corr/abs-2402-03300} provide scalar rewards, DPO~\citep{DBLP:conf/nips/RafailovSMMEF23} is offline and pairwise, and on-policy distillation~\citep{DBLP:journals/corr/abs-2306-13649} uses prompt-level teacher weights, motivating Med-OPD's multimodal structured-CoT training with token-level dual-factor routing. Our formulation is also consistent with systematicity and compositional generalization arguments~\citep{DBLP:journals/corr/abs-2302-01067}, and is orthogonal to backbone-focused medical MLLMs such as RadFM~\citep{wu2025towards}, LLaVA-Med~\citep{DBLP:conf/nips/LiWZULYNPG23}, and MedVInT/PMC-VQA~\citep{DBLP:journals/corr/abs-2305-10415}. VICL~\citep{DBLP:conf/acl/ZhouLWS24} studies task-oriented visual demonstration construction; CADFT~\citep{DBLP:conf/acl/ZhouSWS26} addresses compatibility-aware supervised optimization. CARE~\citep{DBLP:conf/eccv/ZhouLWXS26} and CogAlign~\citep{DBLP:journals/corr/abs-2603-20698} improve medical reasoning through trajectory curation and clinical-cognition alignment, respectively. The full version of related work can be found in Appendix~\ref{app:related}.

\section{Internalized Meta-Capabilities for Agentic Reasoning}
\label{sec:framework}

We now make the paradigm precise. Throughout, $x$ denotes an input
(possibly multimodal), $\mathrm{ctx}$ a context, and $\pi_\theta$ an
autoregressive LLM with parameters $\theta$.
%
%
\begin{definition}[Meta-capability]
\label{def:metacap}
A \emph{meta-capability} of $\pi_\theta$ in an expert domain $\mathcal{D}$
is a function $\kappa$ that, given $x \in \mathcal{D}$ and a context
$\mathrm{ctx}$, produces an output token sequence
$y = \kappa(x, \mathrm{ctx})$ representing one observation about $x$
and satisfying the three properties below.
\end{definition}

\begin{property}[Atomicity]
$\kappa$ corresponds to a single indivisible observational dimension of
$\mathcal{D}$; further decomposition would not yield reusable units.
\end{property}

\begin{property}[Internalization]
$\kappa$ is implemented by the parameters $\theta$ themselves: invocation
$\kappa(x, \mathrm{ctx})$ is a generation step under $\pi_\theta$, not a call
to an external API.
\end{property}

\begin{property}[Compositionality]
A finite set $\{\kappa_1, \ldots, \kappa_K\}$ of meta-capabilities can be
\emph{composed} within a single CoT to solve domain tasks; the composition
policy itself is a function of $(x, \mathrm{ctx}, \theta)$ and is learnable.
\end{property}

\paragraph{Why the three properties are jointly necessary.}
Each missing property corresponds to a failure mode observed in prior
work. Without \emph{atomicity}, capabilities entangle with diseases or
tasks and fail on unseen combinations, as exemplified by end-to-end
specialization. Without \emph{internalization}, capabilities live
behind external runtimes, inheriting their brittleness, latency, and
non-learnability, characteristic of classical agentic tool use.
Without \emph{compositionality}, capabilities exist only in isolation,
requiring hand-designed pipelines to be useful. Internalized
meta-capabilities satisfy all three properties simultaneously.

\subsection{Agentic composition via capability-tagged CoT}
\label{sec:tagged-cot}

We realize agentic composition as a single-pass CoT in which capability
invocations are surfaced as natural-language prefixes. Concretely,
given an input $x$ and an expert query $q$, the model emits a token
sequence of the form
\begin{equation}
\label{eq:traj}
\tau \;=\; \big[\,\langle\textsc{think}\rangle\;\,
\kappa_{i_1}{:}o_1,\;\kappa_{i_2}{:}o_2,\;\ldots,\;\kappa_{i_m}{:}o_m\;\,
\langle/\textsc{think}\rangle\;\,\langle\textsc{answer}\rangle\;d\;\,\langle/\textsc{answer}\rangle\,\big],
\end{equation}
where each $\kappa_{i_j} \in \{\kappa_1,\ldots,\kappa_K\}$ is a
capability key (a plain-text identifier), $o_j$ is its associated
free-form observation, and $d$ is the final answer. The capability
keys are drawn from the model's existing vocabulary, so no token-table
extension is required, and segment boundaries are recovered by a
single regular expression.
By construction, every output token $z_t$ is unambiguously assigned an
active capability tag $c(z_t) \in
\{\kappa_1,\ldots,\kappa_K\}\cup\{\textsc{answer}\}$, which we will
exploit in Section~\ref{sec:medopd} to make the teacher routing depend
on $c(z_t)$. The same token stream therefore encodes both
\emph{which} capability is being invoked and \emph{what} observation
it produces; agentic behavior is realized within a single forward
pass, with neither external runtime nor multi-turn scaffolding.

\paragraph{Generality.}
Although we instantiate the paradigm on medical visual diagnosis, the
three properties are domain-agnostic. Any expert task that can be
reduced to a small set of observational or analytic primitives, such
as legal evidence analysis, mechanical fault diagnosis, security
auditing, or scientific peer review, admits an analogous taxonomy and
an analogous capability-tagged CoT. The remainder of the paper
develops the medical instantiation as a concrete blueprint.

\section{Med-Agentic: Instantiation in Medical Visual Diagnosis}
\label{sec:medagentic}

\subsection{Capability taxonomy and tagged reasoning interface}
\label{sec:taxonomy}

We construct a taxonomy of medical visual meta-capabilities by combining
expert practice, radiological reporting conventions, and the native label
schemata of the $30$ public datasets aggregated in this work
(Section~\ref{sec:exp-setup}). Each capability $\kappa_i$ is required to be
(i) describable in one sentence, (ii) shared across at least three diseases
or anatomies, and (iii) annotatable from existing dataset labels with bounded
effort. The resulting set comprises $K=14$ capabilities, ranging from
low-level visual primitives such as colour and density to higher-level
structural descriptors such as distribution and internal structure, and
including a spatial-grounding capability that localises the region of
interest underlying every observation. The full list, with one-sentence
definitions and per-dataset coverage, is given in
Appendix~\ref{app:taxonomy}.

\begin{wrapfigure}{R}{0.49\linewidth}
  \centering
  \vspace{-4pt}
  \includegraphics[width=\linewidth]{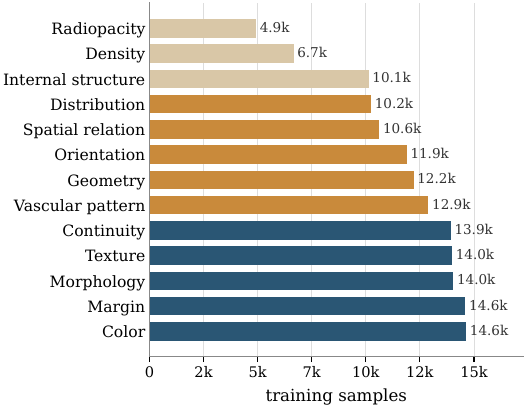}
  \vspace{-17pt}
  \caption{\small Number of training samples per meta-capability ($13$ shown).
  Spatial grounding is not enumerated here because it is exercised by
  every training sample. The distribution is mildly long-tailed, with a
  $3\!:\!1$ ratio between the most and least frequent capability.}
  \label{fig:capfreq}
  \vspace{-7mm}
\end{wrapfigure}

Each training sample carries on average $5$-$6$ capability annotations,
and capability frequencies are mildly long-tailed
(Figure~\ref{fig:capfreq}): the most frequent capability, colour, has
$14{,}638$ training samples, while the rarest, radiopacity, has
$4{,}923$, yielding a roughly
$3\!:\!1$ ratio. We counteract this imbalance by frequency-balanced
sampling in Stage~$1$ of training (Section~\ref{sec:pipeline}). Two
properties of the taxonomy are worth emphasizing. First, capabilities are
defined at the \emph{task level} rather than the disease level: the same
primitive ``irregular margin'' applies to a lung nodule, a skin lesion, and
a thyroid mass, which is what licenses cross-disease reuse. Second, the
$K=14$ axes are not claimed to be unique or exhaustive; we adopt them
because they yield a parsimonious cover of the merged corpus, and we
analyze sensitivity to $K$ empirically by varying it over
$\{2,4,8,14\}$ in Section~\ref{sec:exp}.

\paragraph{Tagged reasoning interface.}
\label{sec:cot}
A Med-Agentic trajectory takes the form of Equation~\ref{eq:traj} with
$K=14$ capabilities, with the per-token tag
$c(z_t) \in \{1,\ldots,14\}\cup\{\textsc{answer}\}$ well-defined
throughout (Section~\ref{sec:tagged-cot}). Thus, the taxonomy specifies
which clinical primitives are available, while the tagged CoT interface
turns their invocation into supervised token segments used by the
training pipeline below.

\subsection{Three-stage training pipeline}
\label{sec:pipeline}

\begin{figure}[t]
  \centering
  \includegraphics[width=\linewidth]{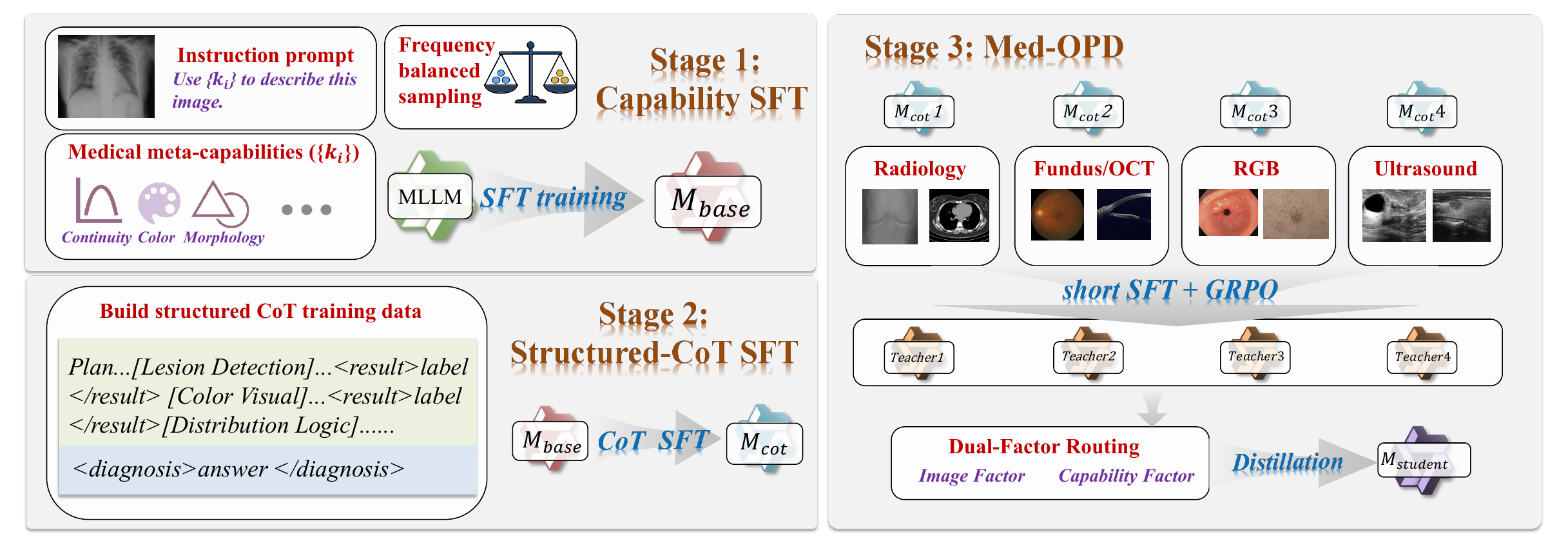}
  \caption{\small Med-Agentic training pipeline. Stages~1-2 produce a cold-start
  $M_{\mathrm{cot}}$ that emits well-formed capability-tagged CoT. Stage~3
  (Med-OPD) is the core methodological contribution.}
  \label{fig:pipeline}
\end{figure}

Med-Agentic is trained in three stages, each with a distinct role
(Figure~\ref{fig:pipeline}).

\paragraph{Stage 1 - Capability SFT.}
We finetune a base MLLM on per-capability instruction data of the form
\texttt{(image, "Use \{$\kappa_i$\} to describe this image.", $\kappa_i$
observation)}, drawing on dataset-derived single-capability annotations.
Capabilities are sampled with frequency-inverse weights so that all
$14$ capabilities are seen approximately equally often. The output,
$M_{\mathrm{base}}$, is fluent on any single capability but does not
yet compose them.

\paragraph{Stage 2 - Structured-CoT SFT.}
We then warm-start composition by finetuning $M_{\mathrm{base}}$ on full
$\langle\textsc{think}\rangle\langle\textsc{answer}\rangle$ trajectories
constructed in two complementary ways: (a) Gemini-3-Flash is prompted with
the capability list and the dataset's native labels to produce a tagged
CoT; (b) given a confirmed diagnosis and the dataset's capability
annotations, we programmatically order the segments along a clinically
plausible sequence (e.g., global-before-local). The output,
$M_{\mathrm{cot}}$, emits well-formed tagged CoT but is not yet a strong
reasoner. Per-stage data, schedules, hyperparameters, and the precise
segment-level loss used in Stage~$2$ are deferred to
Appendices~\ref{app:hparams}-\ref{app:stage12}.

\paragraph{Stage 3 - Med-OPD.}
Finally, we apply the multi-teacher on-policy distillation procedure
of Section~\ref{sec:medopd}.

\subsection{Med-OPD: token-level dual-factor on-policy distillation}
\label{sec:medopd}

Med-OPD distills $N$ modality/anatomy specialists $\{\pi_{T_i}\}_{i=1}^{N}$
($N=4$ in this work) into a single student $\pi_\theta$ that shares the
vocabulary, tokenizer, and vision encoder family. The procedure has two
halves: \emph{specialist forging} and \emph{distillation}.

\paragraph{Specialist forging.}
We partition the $30$ datasets into four clusters based on imaging modality
and anatomy: $T_{\mathrm{radio}}$ (X-ray / CT / MRI), $T_{\mathrm{fundus}}$
(fundus / OCT), $T_{\mathrm{rgb}}$ (endoscopy / dermoscopy / intra-oral
photography), and $T_{\mathrm{us}}$ (ultrasound). Each specialist is
initialized from $M_{\mathrm{cot}}$ and trained only on its cluster, using
a short SFT pass followed by lightweight GRPO~\citep{DBLP:journals/corr/abs-2402-03300} on diagnostic
accuracy. This GRPO step is short and acts only as a per-cluster
calibration of each specialist; it is not the post-training step
itself, which is Med-OPD below. The procedure yields four expert
policies $\pi_{T_1},\ldots,\pi_{T_4}$, each strong on its own cluster
but weak elsewhere.

\paragraph{Token-level dual-factor routing.}
Let $\phi(x)\in\mathbb{R}^{d}$ be the vision-encoder embedding of input $x$,
let $\mu_i \in \mathbb{R}^{d}$ be the centroid of $\phi$ on the training
shard of specialist $i$, and let $\mathbf{W}^{\mathrm{cap}}\in\mathbb{R}_{+}^{N\times(K+1)}$
be a competence matrix whose entry $\mathbf{W}^{\mathrm{cap}}_{i,c}$ records
specialist $i$'s GPT-$4$-as-judge score on capability tag $c$, estimated
once on a held-out probe set and frozen thereafter. We define the
\emph{image factor} as a softmax over negative squared distances,
\begin{equation}
\label{eq:wimg}
w^{\mathrm{img}}_i(x) \;=\; \mathrm{softmax}_i\!\Big(-\lVert \phi(x)-\mu_i\rVert_2^{2}\,/\,\tau_1\Big),
\qquad i=1,\ldots,N,
\end{equation}
and the \emph{capability factor} as a temperature-scaled normalization of
the competence matrix,
\begin{equation}
\label{eq:wcap}
w^{\mathrm{cap}}_i\!\big[c\big] \;=\; \mathrm{softmax}_i\!\Big(\,\log \mathbf{W}^{\mathrm{cap}}_{i,c}\,/\,\tau_2\Big),
\qquad c\in\{1,\ldots,K\}\cup\{\textsc{answer}\}.
\end{equation}
The two factors are combined multiplicatively in log-space to yield the
\emph{token-level dual-factor} routing weight at token $t$,
\begin{equation}
\label{eq:dualfactor}
w_i(t \mid x,\tau) \;=\; \mathrm{softmax}_i\!\Big(\,\log w^{\mathrm{img}}_i(x) + \log w^{\mathrm{cap}}_i\!\big[c(z_t)\big]\,\Big).
\end{equation}
Equation~\ref{eq:dualfactor} is the central object of our method:
it varies both \emph{across inputs} (via the image factor) and
\emph{along the token axis} of a single trajectory (via the capability
factor). Per-prompt static routing, used by recent on-policy
distillation~\citep{DBLP:journals/corr/abs-2306-13649}, is the special case in which $\mathbf{W}^{\mathrm{cap}}$
collapses to a column-constant matrix, so that $w_i(t)\equiv w^{\mathrm{img}}_i(x)$
for every $t$. Conversely, ablating the image factor recovers a
capability-only routing, which we evaluate in Section~\ref{sec:exp}.

\paragraph{Full-vocabulary reverse-KL objective.}
Given $w_i(t)$, the student is updated by minimizing the expected mixture
of token-level reverse Kullback-Leibler divergences against the four
frozen teachers,
\begin{equation}
\label{eq:opd}
\!\!\mathcal{L}(\theta) \!=\!
\mathbb{E}_{(x,q)\sim\mathcal{D}}\;\mathbb{E}_{\tau\sim\pi_\theta(\cdot\mid x,q)}\!
\left[\,\sum_{t=1}^{T}\sum_{i=1}^{N} w_i(t\mid x,\tau)\;
D_{\mathrm{KL}}\!\Big(\,\pi_\theta(\cdot\mid z_{<t},x)\;\Big\|\;\pi_{T_i}(\cdot\mid z_{<t},x)\,\Big)\right],\!\!
\end{equation}
computed over the full vocabulary. For tokens within the answer segment
($c(z_t)=\textsc{answer}$), the capability factor reduces to a uniform
soft mixture over $i$ and the image factor dominates, so the objective
recovers a standard input-conditioned distillation on the final answer.

\paragraph{Theoretical properties.}
Two properties of Equation~\ref{eq:opd} clarify what the dual-factor
routing buys over per-prompt routing.

\begin{proposition}[Strict generalization]
\label{prop:generalize}
Let $\mathcal{L}_{\mathrm{prompt}}(\theta)$ be the per-prompt OPD objective
with weights $\bar w_i(x) \in \Delta^{N-1}$. Then the family of objectives
of the form (\ref{eq:opd}), parameterized by the pair
$(w^{\mathrm{img}}, \mathbf{W}^{\mathrm{cap}})$, strictly contains
$\mathcal{L}_{\mathrm{prompt}}$ as the subfamily obtained by setting
$w^{\mathrm{img}}_i \equiv \bar w_i$ and $\mathbf{W}^{\mathrm{cap}}_{i,c}\equiv 1$
for all $i,c$.
\end{proposition}

\begin{proposition}[Unbiasedness]
\label{prop:unbiased}
Let $g_t(\theta) = \sum_{i} w_i(t\mid x,\tau)\,\nabla_\theta D_{\mathrm{KL}}(\pi_\theta\|\pi_{T_i})|_{z_{<t},x}$
be the per-token gradient under dual-factor routing, and let
$\bar g_t(\theta)$ be the corresponding gradient under per-prompt routing
with weights $\bar w_i(x)=\mathbb{E}_{c\sim p(c\mid x)}[w_i(t\mid x,\tau)]$,
where $p(c\mid x)$ is the empirical capability distribution induced by
$\pi_\theta$. Then the two estimators target the same expected
gradient,
\begin{equation}
\label{eq:unbiased}
\mathbb{E}[g_t(\theta)\mid x] \;=\; \mathbb{E}[\bar g_t(\theta)\mid x],
\end{equation}
so dual-factor routing introduces no bias relative to per-prompt
routing despite reweighting along the token axis.
\end{proposition}

\noindent\emph{Sketch.} Taking expectation of $g_t(\theta)$ over the
capability random variable $c(z_t)$ given $x$ yields
$\mathbb{E}_{c\mid x}[w_i(t\mid x,\tau)] = \bar w_i(x)$, and pulling the
expectation through the linear sum recovers $\bar g_t(\theta)$. A full
proof and the precise regularity conditions are deferred to
Appendix~\ref{app:proofs}.

\paragraph{Why this matters in medical reasoning.}
Per-prompt routing is appropriate when each prompt belongs to a single
domain, for example math versus code. In medical reasoning, by contrast, a
single trajectory $\tau$ typically spans multiple capabilities, and the
best specialist depends on \emph{which} capability the current token is
expressing: a fundus specialist is most informative when the student is
generating a \texttt{vascular\_pattern} segment on a fundus image, but a
radiology specialist is more informative when the same trajectory enters
the radiopacity capability. Equation~\ref{eq:dualfactor} captures
this dependency exactly, while remaining negligible in compute: both
factors are scalars precomputed at coarser granularities and combined once
per token.

\paragraph{Implementation.}
Vision-encoder consistency, teacher offloading, the held-out probe
set used to estimate $\mathbf{W}^{\mathrm{cap}}$, and the
inter-annotator agreement audit are detailed in
Appendix~\ref{app:impl}.

\section{Experiments}
\label{sec:exp}

We evaluate Med-Agentic on ID and OOD medical images and dissect the
three-stage pipeline (Section~\ref{sec:exp-main}), attribute the gain
to each of the three properties of an internalized meta-capability
(Section~\ref{sec:exp-properties}), study the Med-OPD routing mechanism
(Section~\ref{sec:exp-opd}), and inspect the interpretability of the
learned capability policy (Section~\ref{sec:exp-interp}). Further
ablations on Stage-1 sampling, capability granularity, the CoT
interface, the specialist count $N$, and system-level consequences of
internalization are deferred to Appendix~\ref{app:extra-ablations}.

\subsection{Setup}
\label{sec:exp-setup}

\emph{Data.} We aggregate $30$ public medical imaging datasets
($9$ anatomies, $8$ modalities, $119{,}373$ samples) into a
$29{,}992$-sample training pool, a $6{,}033$-sample ID test set, and a
$519$-sample OOD set spanning three held-out slices
(\emph{Head-CT hemorrhage}, \emph{COVID-19 CT},
\emph{HyperKvasir held-out grades}) that probe new modality/anatomy,
cross-modality cross-disease, and label-shift generalization. The
remaining $82{,}829$ samples are held back for future scaling studies.
Each capability is supported by $5$--$21$ source datasets, except
spatial grounding which is exercised by every training sample.
Per-dataset and per-capability details are in
Appendix~\ref{app:datasets}.

\emph{Baselines.} All trained methods share the same Qwen3.5-9B
backbone~\citep{DBLP:journals/corr/abs-2502-13923}, so gaps reflect the training recipe rather than scale.
We compare against \emph{(i)} closed-source MLLMs
(Gemini-2.5/3.0-Flash~\citep{DBLP:journals/corr/abs-2312-11805}, GPT-5.5~\citep{DBLP:journals/corr/abs-2303-08774}, Claude Sonnet 4.6~\citep{DBLP:journals/corr/abs-2212-08073}),
\emph{(ii)} open general MLLMs (Qwen3.5-9B/27B~\citep{DBLP:journals/corr/abs-2502-13923}),
\emph{(iii)} open medical MLLMs (Hulu-Med-7B~\citep{DBLP:journals/corr/abs-2510-08668}, HuatuoGPT-V~\citep{DBLP:journals/corr/abs-2406-19280}),
and \emph{(iv)} four trained controls that each ablate one component:
disease-level multi-task SFT (removes atomicity), a fixed pipeline of
capability tools (removes compositionality), external tool
orchestration over the same Stage-3 specialists (removes
internalization), and plain GRPO~\citep{DBLP:journals/corr/abs-2402-03300} from $M_{\mathrm{cot}}$ with
accuracy reward (removes Med-OPD). The full roster of twelve methods
is in Appendix~\ref{app:methods}.

\emph{Metrics.} We report top-$1$ accuracy, macro-$F_1$, the fraction
of CoT trajectories that pass structural parsing (``CoT WF''), and ROI
intersection-over-union (``IoU''). On OOD we additionally report
$\Delta$top-$1$ vs.\ the Qwen3.5-9B zero-shot control to net out
language priors. CoT quality is judged by GPT-$4$ along faithfulness,
ordering, and specificity, with a $100$-sample human spot-check
(Appendix~\ref{app:annotation}). All numbers are means over three
seeds; closed- and open-source rivals are evaluated zero-shot.

\emph{Implementation.} All trained methods are run on a single
$8\times$H200 node in bf16 with FSDP and FlashAttention-$2$, sharing
the Qwen3.5-9B backbone, tokenizer, and vision encoder.
Synthetic capability and trajectory annotations are produced by
Gemini-$3$-Flash through the seven prompts of
Appendix~\ref{app:prompts}; the annotator stack, scale, filtering,
and human spot-check are detailed in Appendix~\ref{app:syndata}; the
per-stage hyperparameters are in Appendix~\ref{app:hparams}; the
precise Stage~$1$ and Stage~$2$ objectives, including the
segment-level loss, are in Appendix~\ref{app:stage12}.

\subsection{In-distribution and out-of-distribution results}
\label{sec:exp-main}

\paragraph{Accuracy.}
\begin{wraptable}[20]{R}{0.48\linewidth}
\vspace{-9mm}
\centering
\setlength{\tabcolsep}{4pt}
\caption{\small ID test results on $n = 6{,}033$ samples. ``CoT WF'' is the
fraction of CoT trajectories that pass the structural parser;
\textsf{n/a} indicates that the method does not emit tagged CoT.
$^{\dagger}$~Hand-wired templates trivially pass the regex.}
\label{tab:exp-id}
\resizebox{\linewidth}{!}{
\begin{tabular}{@{}lcccc@{}}
\toprule
Method & Top-$1$ & $F_1$ & CoT WF & IoU \\
\midrule
Gemini-2.5-Flash~\citep{DBLP:journals/corr/abs-2312-11805}   & 17.4 & 14.9 & 41.2 & 0.105 \\
Gemini-3.0-Flash~\citep{DBLP:journals/corr/abs-2312-11805}   & 22.9 & 19.6 & 53.7 & 0.142 \\
GPT-5.5~\citep{DBLP:journals/corr/abs-2303-08774}             & 28.3 & 24.7 & 64.8 & 0.187 \\
Claude Sonnet 4.6~\citep{DBLP:journals/corr/abs-2212-08073}  & 26.5 & 23.0 & 61.4 & 0.176 \\
Qwen3.5-9B~\citep{DBLP:journals/corr/abs-2502-13923}          & 13.8 & 11.2 & 28.5 & 0.082 \\
Qwen3.5-27B~\citep{DBLP:journals/corr/abs-2502-13923}         & 19.2 & 16.4 & 38.7 & 0.118 \\
Hulu-Med-7B~\citep{DBLP:journals/corr/abs-2510-08668}         & 11.6 &  9.4 & 32.4 & 0.094 \\
HuatuoGPT-V~\citep{DBLP:journals/corr/abs-2406-19280}         & 14.7 & 12.3 & 35.0 & 0.106 \\
\midrule
Disease-level SFT         & 60.1 & 56.4 & \textsf{n/a}      & 0.286 \\
Fixed pipeline             & 62.5 & 58.9 & $100^{\dagger}$    & 0.378 \\
External tool              & 63.4 & 59.7 & 87.6               & 0.394 \\
Plain GRPO~\citep{DBLP:journals/corr/abs-2402-03300}          & 65.8 & 62.3 & 95.2               & 0.402 \\
\textbf{Ours}             & \textbf{69.7} & \textbf{66.4} & \textbf{96.8} & \textbf{0.431} \\
\bottomrule
\end{tabular}}
\vspace{-10mm}
\end{wraptable}
Tables~\ref{tab:exp-id}--\ref{tab:exp-ood} report ID and OOD
performance. Closed-source general MLLMs solve at most $28.3\%$ of the
$6{,}033$ in-distribution cases zero-shot; the medically specialized
Hulu-Med-7B trails further behind ($11.6\%$). Every backbone trained on
our $29{,}992$-sample pool clears $60\%$, confirming that domain
supervision dominates the zero-shot baselines. Within the trained
family, Med-Agentic delivers $\mathbf{69.7\%}$ top-$1$ and
$\mathbf{66.4\%}$ macro-$F_1$, a $+3.9$-point gain over plain GRPO, the
strongest trained control, and a $+41.4$-point gain over GPT-5.5. Its
CoT well-formed rate of $96.8\%$ matches the human-wired fixed
pipeline ($100\%$ by construction), while its grounding IoU of
$0.431$ is the highest across the board.

On OOD, the three slices probe different kinds of distribution shift:
COVID-19 CT is a cross-modality/cross-disease shift within a shared
anatomy; HyperKvasir-OOD is a label shift on held-out severity grades;
and Head-CT hemorrhage is a cross-modality, emergent-disease shift on a
rare anatomy. Zero-shot closed-source MLLMs cluster in the
$10$--$25\%$ range across all three. Among the trained controls, the
ranking tracks how many of the three paradigm properties are preserved:
disease-level SFT ($35.8$ avg.) $<$ fixed pipeline ($38.8$) $<$ external
tool ($43.3$) $<$ plain GRPO ($50.3$) $<$ \textbf{Ours} ($56.5$). The
$+6.2$-point gap of Ours over plain GRPO, combined with the
$+13.2$-point gap over external-tool orchestration, isolates the
contribution of Med-OPD from that of generic on-policy post-training
and from that of mere tool access.

\begin{table}[t]\small
\centering
\caption{\small OOD zero-shot top-$1$ on three held-out slices. For trained
methods the subscript in \subgain{+XX.X} is the gain over the
Qwen3.5-9B zero-shot control; zero-shot rivals use a different backbone
and are reported as a bare number. ``Avg.'' is the unweighted mean over
the three slices.}
\label{tab:exp-ood}
\small
\setlength{\tabcolsep}{4pt}
\begin{tabular}{@{}lcccc@{}}
\toprule
Method & COVID-19 CT & HyperKvasir-OOD & Head-CT hem. & Avg. \\
\midrule
Gemini-2.5-Flash~\citep{DBLP:journals/corr/abs-2312-11805}    & 12.9 & 16.5 &  9.0 & 12.8 \\
Gemini-3.0-Flash~\citep{DBLP:journals/corr/abs-2312-11805}    & 18.1 & 22.0 & 14.5 & 18.2 \\
GPT-5.5~\citep{DBLP:journals/corr/abs-2303-08774}                    & 23.3 & 27.5 & 18.6 & 23.1 \\
Claude Sonnet 4.6~\citep{DBLP:journals/corr/abs-2212-08073}    & 21.5 & 25.4 & 17.0 & 21.3 \\
Qwen3.5-9B~\citep{DBLP:journals/corr/abs-2502-13923}                       &  8.6 & 11.0 &  5.4 &  8.3 \\
Qwen3.5-27B~\citep{DBLP:journals/corr/abs-2502-13923}                      & 14.6 & 17.5 &  9.7 & 13.9 \\
Hulu-Med-7B~\citep{DBLP:journals/corr/abs-2510-08668}                         &  7.4 &  9.8 &  4.6 &  7.3 \\
HuatuoGPT-V~\citep{DBLP:journals/corr/abs-2406-19280}                    & 10.0 & 12.8 &  6.3 &  9.7 \\
\midrule
Disease-level SFT          & 38.7\subgain{+30.1} & 47.2\subgain{+36.2} & 21.4\subgain{+16.0} & 35.8\subgain{+27.5} \\
Fixed pipeline             & 41.4\subgain{+32.8} & 50.3\subgain{+39.3} & 24.7\subgain{+19.3} & 38.8\subgain{+30.5} \\
External tool              & 47.2\subgain{+38.6} & 54.6\subgain{+43.6} & 28.0\subgain{+22.6} & 43.3\subgain{+35.0} \\
Plain GRPO~\citep{DBLP:journals/corr/abs-2402-03300}            & 56.0\subgain{+47.4} & 61.5\subgain{+50.5} & 33.4\subgain{+28.0} & 50.3\subgain{+42.0} \\
\textbf{Ours}              & \textbf{62.7}\subgain{+54.1} & \textbf{67.8}\subgain{+56.8} & \textbf{39.2}\subgain{+33.8} & \textbf{56.5}\subgain{+48.2} \\
\bottomrule
\end{tabular}
\end{table}

\begin{figure}[t]
  \centering
  \begin{minipage}[t]{0.44\linewidth}
    \centering
    \includegraphics[width=\linewidth]{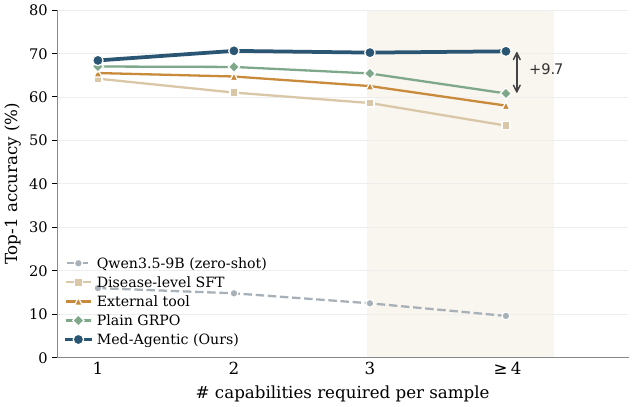}
  \end{minipage}
  \hfill
  \begin{minipage}[t]{0.54\linewidth}
    \centering
    \includegraphics[width=\linewidth]{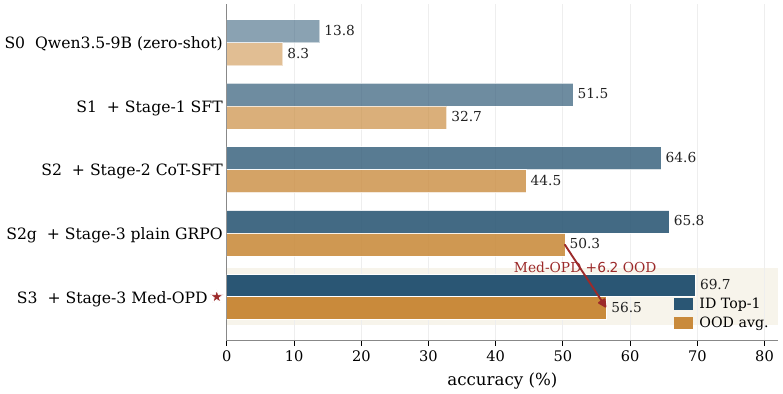}
  \end{minipage}
  \caption{\small Compositional generalization and pipeline dissection.
  \textbf{Left:} Top-$1$ accuracy as a function of the number of
  capabilities required per sample. Med-Agentic is nearly flat, while
  every other trained baseline drops by at least $7$ points on the
  high-complexity bucket (shaded).
  \textbf{Right:} Three-stage pipeline ablation; paired bars show ID
  top-$1$ (teal) and OOD average (ochre) for each variant, and the
  arrow marks the $+6.2$-point OOD gain of Med-OPD over plain GRPO.}
  \label{fig:exp-main-ablations}
\end{figure}

\paragraph{Compositional-complexity slicing.}
To make the compositionality claim quantitative, we partition the ID
and OOD sets by the \emph{number of capabilities required} by the
ground-truth CoT and evaluate each method on the resulting buckets
(left panel of Figure~\ref{fig:exp-main-ablations}). All trained
baselines are within $3$ points of one another on single-capability
questions, but they fan out as complexity grows: on samples that
require at least four capabilities, plain GRPO drops by $6$ points,
external-tool orchestration by $7.5$, and disease-level SFT by $11$,
while Med-Agentic stays within $\pm 1$ point of its single-capability
performance ($68.4$ to $70.5$ across all buckets). This is direct
evidence that internalized, agentically composed capabilities
generalize on composition depth, whereas monolithic specialists do
not.

\paragraph{Pipeline dissection.}
The right panel of Figure~\ref{fig:exp-main-ablations} dissects the
three-stage recipe of Section~\ref{sec:pipeline} into five variants:
\textbf{S0} the zero-shot Qwen3.5-9B backbone, \textbf{S1} Stage-1
only, \textbf{S2} Stages~1--2, \textbf{S2g} Stages~1--2 with plain GRPO
substituted for Stage~3, and \textbf{S3} the full Med-OPD pipeline.
Stage~1 alone lifts zero-shot Qwen3.5-9B from $13.8 / 8.3$ to $51.5 /
32.7$ (ID/OOD), the regime in which the student understands individual
capabilities. Stage~2 (structured-CoT SFT) adds $+13.1 / +11.8$ by
teaching capability-segment composition. Replacing Stage~3 with plain
GRPO (S2g) gains $+5.8$ on OOD but only $+1.2$ on ID; substituting
Med-OPD (S3) adds $+3.9 / +6.2$ on top of the GRPO variant. Med-OPD
thus converts the on-policy reward signal into a multi-teacher
distillation signal that preserves OOD robustness without compromising
ID fidelity.

\subsection{Necessity of the three properties}
\label{sec:exp-properties}
\begin{table}[!t]
\centering
\caption{\small Necessity of the three properties. Each row drops exactly one
property relative to Ours. ID and OOD are top-$1$ in percent;
$\Delta$OOD is the OOD drop relative to row P0.}
\label{tab:exp-properties}
\small
\begin{tabular}{@{}llcccccc@{}}
\toprule
    & Variant                                       & Atom. & Intern. & Comp. & ID & OOD & $\Delta$OOD \\
\midrule
P0  & Med-Agentic (full)                            & \checkmark & \checkmark & \checkmark & \textbf{69.7} & \textbf{56.5} & --- \\
P1  & $-$ atomicity (disease-level SFT)             & \ding{55}  & \checkmark & \checkmark & 60.1          & 35.8          & $-20.7$ \\
P2  & $-$ internalization (external tools)          & \checkmark & \ding{55}  & \checkmark & 63.4          & 43.3          & $-13.2$ \\
P3  & $-$ compositionality (fixed pipeline)         & \checkmark & \checkmark & \ding{55}  & 62.5          & 38.8          & $-17.7$ \\
\bottomrule
\end{tabular}
\end{table}
Section~\ref{sec:framework} isolates atomicity, internalization, and
compositionality as the three defining properties of an internalized
meta-capability. Table~\ref{tab:exp-properties} confirms that removing
any one of them costs substantial OOD accuracy, while ID degrades much
less. Atomicity is the most load-bearing: disease-level SFT loses
$20.7$ points on OOD because it memorizes disease-to-label mappings
rather than reusing basic visual observations. Internalization and
compositionality each cost $13.2$ and $17.7$ points on OOD
respectively, and each failure mode is consistent with the mechanism
identified in Section~\ref{sec:framework}, namely error propagation
through brittle tool interfaces, and the inability to reorder
capability segments on unseen diseases.

\subsection{CoT interface and Med-OPD post-training}
\label{sec:exp-opd}

\paragraph{Med-OPD routing.}
Table~\ref{tab:exp-routing} ablates the two routing factors of Med-OPD.
Uniform teacher mixing (R0) yields $65.6 / 50.4$ on ID/OOD.
Per-prompt image routing (R1), which corresponds to the degenerate
case of Proposition~\ref{prop:generalize} with $w^{\mathrm{cap}}$
constant, adds $+1.5 / +2.3$. Capability-only routing (R2) is slightly
stronger. The full dual-factor router (R3) strictly subsumes both and
attains $69.7 / 56.5$, i.e.\ $+2.6 / +3.8$ over the strongest
degenerate case. By Proposition~\ref{prop:unbiased}, R3 is an
unbiased estimator of the per-prompt expected gradient under matched
marginals; the empirical gain of R3 over R1 is therefore most
naturally explained by the strictly larger hypothesis class
(Proposition~\ref{prop:generalize}) rather than by a biased estimator
target.
\begin{wraptable}{R}{0.40\linewidth}
\vspace{-8mm}
\centering
\small
\setlength{\tabcolsep}{3pt}
\caption{\small Med-OPD routing ablation.
$w^{\mathrm{i}}\!\equiv\!w^{\mathrm{img}}$ (Eq.~\ref{eq:wimg});
$w^{\mathrm{c}}\!\equiv\!w^{\mathrm{cap}}$ (Eq.~\ref{eq:wcap}).
ID/OOD are top-$1$.}
\label{tab:exp-routing}
\begin{tabular}{@{}llccrr@{}}
\toprule
    & Routing                       & $w^{\mathrm{i}}$ & $w^{\mathrm{c}}$ & ID & OOD \\
\midrule
R0  & Uniform                        & ---        & ---        & 65.6          & 50.4 \\
R1  & Image-only                     & \checkmark & ---        & 67.1          & 52.7 \\
R2  & Capability-only                & ---        & \checkmark & 67.8          & 53.9 \\
R3  & \textbf{Dual-factor}           & \checkmark & \checkmark & \textbf{69.7} & \textbf{56.5} \\
\bottomrule
\end{tabular}
\vspace{-3em}
\end{wraptable}

\paragraph{Further ablations.}
Capability granularity ($K \in \{2,4,8,14\}$, monotonic gains
supporting $K=14$), CoT-interface design (segment-level loss adds
$+2.4$ top-$1$ and lifts CoT well-formedness to $96.8\%$),
specialist count ($N=4$ peaks; $N=8$ thins per-specialist data and
loses $-2.5$ OOD), and the system-level consequences of internalization
($18.4\% \to 4.0\%$ out-of-$\mathcal{K}^{\star}$ call rate,
$4.6\times$ reduction) are deferred to
Appendix~\ref{app:extra-ablations}.

\subsection{Interpretability of the learned capability policy}
\label{sec:exp-interp}

\begin{figure}[t]
  \centering
  \includegraphics[width=\linewidth]{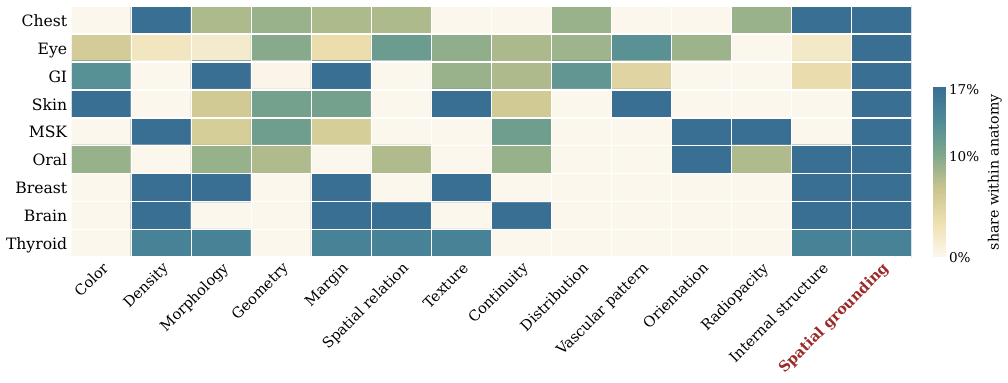}
  \caption{\small Capability-by-anatomy usage heat map (column-normalized).
  Spatial grounding (rightmost) is exercised by every image; the rest
  of the profile tracks radiology practice
  ($\rho \approx 0.78$ against a clinician-authored SOP).}
  \label{fig:exp-heatmap}
\end{figure}

Figure~\ref{fig:exp-heatmap} plots the column-normalized capability
share for each anatomy, aggregated over the Stage-3 student's
trajectories on the ID test set. The emergent specialization aligns
with radiology practice: chest and brain images rely heavily on
density, radiopacity, and internal-structure; eye, skin, and GI
images, which constitute the RGB and fundus families, are dominated by
colour, margin, and vascular-pattern observations; and spatial
grounding is exercised uniformly across all anatomies. The Spearman
rank correlation between the learned capability order and a
clinician-authored SOP reaches $\rho \approx 0.78$ ($p < 0.001$). The
same capability vocabulary is reused on OOD: the per-capability usage
distribution shifts only mildly,
$\mathrm{KL}(P_{\mathrm{OOD}} \,\|\, P_{\mathrm{ID}}) \approx 0.11$~nats,
and reweights in physically coherent directions (CT-heavy slices raise
density, radiopacity, and internal-structure; RGB slices raise colour
and vascular pattern); the per-slice histograms are deferred to
Appendix~\ref{app:extra-ood-usage}.

\paragraph{Additional validation.}
Diagnosis-withheld supervision, capability-segment interventions, and combination-disjoint tests further examine the reasoning interface (Appendix~\ref{app:validation}). Fully withholding diagnosis information gives $67.2/52.9$ ID/OOD accuracy; deleting a relevant observation reduces ID accuracy to $61.8\%$, versus $68.4\%$ for an irrelevant deletion. Accuracy on unseen capability pairs/triples is $65.1/62.4\%$. PCam and guideline-retrieval tests examine transfer limits and complementary external knowledge. These controls support answer dependence and recombination within the fixed basis, rather than independent neural modules or unrestricted transfer. Appendix~\ref{app:eval-protocol} separately reports closed-set re-evaluation of the zero-shot models and clarified data accounting; these results use a different protocol from the original comparisons above.

\section{Conclusion}
\label{sec:conclusion}

We introduced \emph{internalized meta-capabilities}, atomic,
model-internal, and agentically composable reasoning units, as a
third paradigm between external agentic tool use and monolithic
end-to-end specialization, and instantiated it as \textbf{Med-Agentic}
with \textbf{Med-OPD}, a multi-teacher on-policy distillation recipe
whose token-level dual-factor routing strictly generalizes per-prompt
static routing without changing the expected teacher-KL gradient.
Across $30$ medical imaging datasets Med-Agentic improves top-$1$
accuracy over the strongest trained baseline by $3.9$ points on
in-distribution test and $6.2$ points on held-out OOD slices, and
reuses a consistent capability vocabulary under distribution shift;
we expect the taxonomy, capability-tagged CoT interface, and
dual-factor post-training recipe to transfer to other expert domains
whose expertise is decomposable into a small set of reusable
observations.

\bibliographystyle{plain}
\bibliography{ref_camera_ready}
\clearpage
\appendix

\section{Related Work}
\label{app:related}
ReAct~\citep{DBLP:conf/iclr/YaoZYDSN023}, Toolformer~\citep{DBLP:conf/nips/SchickDDRLHZCS23}, ARTIST~\citep{DBLP:journals/corr/abs-2505-01441}, ReTool~\citep{DBLP:journals/corr/abs-2504-11536}, and
related systems formulate expert behavior as the orchestration of
external tools by an LLM planner; in medicine, Octopus~\citep{DBLP:journals/corr/abs-2511-15351} and
MedAgent-Pro~\citep{DBLP:journals/corr/abs-2503-18968} instantiate this pattern with imaging and
report-writing modules. These systems are
agentic in the multi-turn sense, but their capabilities live outside the
model and depend on a runtime that the model itself cannot improve through
training. Med-Agentic shares the agentic spirit but \emph{internalizes}
capabilities into the LLM's parameters, so that ``calling a capability''
reduces to generating a tagged CoT segment.
Recent work attempts to push specialized abilities into LLM parameters.
OneThinker~\citep{DBLP:journals/corr/abs-2512-03043} unifies multiple visual tasks via SFT plus EMA-GRPO, but
treats each task as a single coarse skill and does not expose a compositional
interface. Skill0~\citep{lu2026skill0} internalizes in-context skills via reinforcement
learning, again at task-level granularity and without a structured CoT or
multimodal evaluation. Both demonstrate the value of internalization but
stop short of an \emph{atomic, composable} formulation. Med-Agentic differs
in two ways: capabilities are deliberately atomic, each capturing one
indivisible observational dimension shared across diseases; and the
composition of capabilities is itself a learned, token-level policy.

RLHF~\citep{DBLP:conf/nips/Ouyang0JAWMZASR22} and GRPO~\citep{DBLP:journals/corr/abs-2402-03300} optimize a single scalar reward and
therefore provide a coarse gradient signal for the multi-capability
trajectories considered here; DPO~\citep{DBLP:conf/nips/RafailovSMMEF23} avoids reward modeling at the
cost of being offline and pairwise. Closest to us, recent on-policy
distillation~\citep{DBLP:journals/corr/abs-2306-13649} performs full-vocabulary reverse-KL distillation from
multiple teachers into a student, with \emph{per-prompt} static teacher
weights, which are adequate when each prompt belongs to a single domain (e.g., math
versus code). Med-OPD extends this line along three orthogonal axes: it
operates on multimodal inputs, on structured-CoT trajectories rather than
free-form text, and most importantly with a \emph{token-level dual-factor}
routing whose weights vary across the tokens of a single trajectory.
PF-OPSD~\citep{DBLP:journals/corr/abs-2606-03603} uses teacher-side privileged future information for on-policy self-distillation of concrete visual reasoning. Med-OPD instead routes among modality/anatomy specialists using both the current image and active capability tag.
The cognitive science literature on systematicity and recent
machine-learning analyses of compositional generalization~\citep{DBLP:journals/corr/abs-2302-01067} argue
that generalization to novel combinations arises from systematic
recombination of primitives. Expert clinical reasoning is a textbook
instance: the same set of visual primitives is recombined to
characterize many diseases. Med-Agentic makes this hypothesis testable
by exposing capability invocation as a token stream, enabling
controlled studies of how primitive count interacts with generalization.
RadFM~\citep{wu2025towards}, LLaVA-Med~\citep{DBLP:conf/nips/LiWZULYNPG23}, MedVInT/PMC-VQA~\citep{DBLP:journals/corr/abs-2305-10415}, and related medical
MLLMs target specific imaging modalities or instruction-following
style. Our work is orthogonal: we propose a paradigm and a training
procedure rather than a new backbone, and any of these models can in
principle be repurposed as the student or specialist within
Med-Agentic.

CARE~\citep{DBLP:conf/eccv/ZhouLWXS26} curates medical reasoning trajectories for causal consistency and learnability, while CogAlign~\citep{DBLP:journals/corr/abs-2603-20698} internalizes hierarchical clinical cognition through SFT and counterfactual reinforcement learning. Our focus is a cross-disease atomic capability interface and token-level specialist distillation.

\section{Method roster}
\label{app:methods}

Table~\ref{tab:app-methods} summarizes the twelve methods evaluated in
Section~\ref{sec:exp}. Zero-shot rivals are grouped into closed-source
general MLLMs, open-source general MLLMs, and open-source medical
MLLMs. Trained controls all share the Qwen3.5-9B backbone with
Med-Agentic and differ from it in exactly one design dimension, so
that each controls the contribution of a single paradigm property
(atomicity, internalization, compositionality) or of the Med-OPD
post-training step.

\begin{table}[ht]
\centering
\caption{\small Methods evaluated in Section~\ref{sec:exp}. Trained controls
share the Qwen3.5-9B backbone and the $29{,}992$-sample training pool
with Med-Agentic. ``Component ablated'' indicates which design
dimension is removed relative to our full method.}
\label{tab:app-methods}
\small
\begin{tabular}{@{}lll@{}}
\toprule
Method & Family & Role / component ablated \\
\midrule
Gemini-2.5-Flash~\citep{DBLP:journals/corr/abs-2312-11805}    & Closed general MLLM & Zero-shot \\
Gemini-3.0-Flash~\citep{DBLP:journals/corr/abs-2312-11805}    & Closed general MLLM & Zero-shot \\
GPT-5.5~\citep{DBLP:journals/corr/abs-2303-08774}                    & Closed general MLLM & Zero-shot \\
Claude Sonnet 4.6~\citep{DBLP:journals/corr/abs-2212-08073}    & Closed general MLLM & Zero-shot \\
Qwen3.5-9B~\citep{DBLP:journals/corr/abs-2502-13923}                       & Open general MLLM   & Zero-shot \\
Qwen3.5-27B~\citep{DBLP:journals/corr/abs-2502-13923}                      & Open general MLLM   & Zero-shot \\
Hulu-Med-7B~\citep{DBLP:journals/corr/abs-2510-08668}                         & Open medical MLLM   & Zero-shot \\
HuatuoGPT-V~\citep{DBLP:journals/corr/abs-2406-19280}                    & Open medical MLLM   & Zero-shot \\
\midrule
Disease-level multi-task SFT     & Trained control & $-$ atomicity \\
Fixed pipeline                    & Trained control & $-$ compositionality \\
External tool orchestration       & Trained control & $-$ internalization \\
Plain GRPO~\citep{DBLP:journals/corr/abs-2402-03300}            & Trained control & $-$ Med-OPD post-training \\
\textbf{Med-Agentic (Ours)}       & Full method      & --- \\
\bottomrule
\end{tabular}
\end{table}

\section{Full capability taxonomy}
\label{app:taxonomy}

Table~\ref{tab:taxonomy} lists all $K=14$ medical visual meta-capabilities
together with one-sentence descriptions. Each $\kappa_i$ is task-level: the
same primitive is reused across diseases, modalities, and anatomies.
Per-capability frequencies in the training set are visualized in
Figure~\ref{fig:capfreq} of the main text, with the exception of spatial
grounding, which is exercised by every sample and therefore omitted from
that plot.

\begin{table}[ht]
\centering
\small
\caption{\small The $14$ medical visual meta-capabilities used in Med-Agentic.}
\label{tab:taxonomy}
\begin{tabular}{@{}rll@{}}
\toprule
\# & Capability & One-sentence description \\
\midrule
1  & Color                & color, pigmentation, hemorrhagic hue, and other chromatic cues \\
2  & Density              & CT Hounsfield units, MRI signal intensity, ultrasound echogenicity \\
3  & Morphology           & geometric contour, lobulation, spiculation, overall shape \\
4  & Geometry             & lengths, ratios (e.g., VCDR), angles, areas, metric measurements \\
5  & Margin               & sharpness, infiltration, halo, boundary irregularity \\
6  & Spatial relation     & distance to landmarks, layered relations, anatomical localization \\
7  & Texture              & homogeneity, microstructure, surface texture \\
8  & Continuity           & cortical / tissue-layer / vascular continuity and disruptions \\
9  & Distribution         & scattered vs.\ clustered, symmetry, quadrant-level distribution \\
10 & Vascular pattern     & vessel course, density, neovascularization, occlusion \\
11 & Orientation          & axial alignment, joint apposition, anatomical orientation \\
12 & Radiopacity          & radiopaque vs.\ radiolucent characteristics on radiographs \\
13 & Internal structure   & cavitation, calcification, septation, internal substructure \\
\midrule
14 & Spatial grounding    & localizes the region of interest underlying every attribute judgment \\
\bottomrule
\end{tabular}
\end{table}

\section{Source datasets and per-capability coverage}
\label{app:datasets}

We aggregate $30$ publicly available medical imaging datasets spanning
$9$ anatomical sites (chest, eye, gastrointestinal, skin, musculoskeletal,
oral, breast, brain, thyroid) and $8$ imaging modalities (X-ray, CT, MRI,
fundus, OCT, RGB endoscopy / dermoscopy / intra-oral, ultrasound, panoramic
dental). The merged corpus contains $119{,}373$ unique samples,
partitioned into a training pool of $29{,}992$, an in-distribution test
set of $6{,}033$, an OOD analysis set of $519$, and a reserve of $82{,}829$
samples held back for future scaling studies. Each capability is
supported by $5$-$21$ source datasets, with per-capability training
counts ranging from $4{,}923$ (radiopacity) to $14{,}638$ (colour);
spatial grounding is exercised by every training sample and is therefore
not separately enumerated.

\begin{table}[ht]
\centering
\small
\caption{\small Source datasets aggregated in Med-Agentic. ``Train'' is the
training pool, ``ID'' is the in-distribution test set, and ``OOD'' is the
out-of-distribution analysis slice; the remainder of each public split
is held in reserve.}
\label{tab:datasetlist}
\begin{tabular}{@{}llrrr@{}}
\toprule
Dataset & Modality / anatomy & Train & ID & OOD \\
\midrule
Chest X-ray Pneumonia            & X-ray / chest         &   409 &  200 &   0 \\
COVID-19 CT                      & CT / chest            &     0 &    0 & 116 \\
Chest CT (lung-cancer subtypes)  & CT / chest            &   372 &   93 &   0 \\
IDRiD                            & Fundus / retina       &   366 &   91 &   0 \\
Retinal-Lesions                  & Fundus / retina       &   529 &  100 &   0 \\
PALM (pathologic myopia)         & Fundus / retina       &   338 &   84 &   0 \\
AGE (AS-OCT angle closure)       & OCT / anterior segment&   800 &  200 &   0 \\
ADAM (AMD detection)             & Fundus / retina       &   286 &   70 &   0 \\
REFUGE2 (glaucoma)               & Fundus / retina       &   256 &   64 &   0 \\
GAMMA (glaucoma multi-modal)     & Fundus + OCT / retina &   160 &   40 &   0 \\
GOALS (OCT layer segmentation)   & OCT / retina          &   160 &   40 &   0 \\
ODIR-5K                          & Fundus / retina       & 2{,}303 &  414 &   0 \\
MuReD (multilabel retinal)       & Fundus / retina       & 1{,}303 &  324 &   0 \\
RFMiD                            & Fundus / retina       & 1{,}920 &  482 &   0 \\
BiDR (bilateral DR)              & Fundus / retina       &   800 &  200 &   0 \\
AMD-SD (Spectralis OCT)          & OCT / retina          &   400 &  100 &   0 \\
PolypGen                         & RGB endoscopy / GI    &    30 &    0 &   0 \\
Kvasir V2                        & RGB endoscopy / GI    & 3{,}200 &    0 &   0 \\
Kvasir-SEG                       & RGB endoscopy / GI    &    29 &    0 &   0 \\
HyperKvasir                      & RGB endoscopy / GI    & 1{,}601 &  305 & 200 \\
Kvasir-SEG (lesion seg.)         & RGB endoscopy / GI    & 1{,}519 &  213 &   0 \\
CP-CHILD                         & RGB endoscopy / GI    &   415 &  125 &   0 \\
HyperKvasir-seg                  & RGB endoscopy / GI    &    26 &    0 &   0 \\
ISIC-MILK10k                     & Dermoscopy / skin     & 2{,}400 &  600 &   0 \\
PAD-UFES-20                      & Smartphone / skin     & 1{,}338 &  333 &   0 \\
FracAtlas                        & X-ray / MSK           &   800 &  100 &   0 \\
Digital Knee X-ray (KL grading)  & X-ray / MSK           &   800 &  100 &   0 \\
Knee-OA-Severity (KL grading)    & X-ray / MSK           &   800 &  200 &   0 \\
Oral Diseases                    & Intra-oral RGB / oral & 2{,}400 &  600 &   0 \\
Dental Panoramic Radiography     & X-ray / oral          & 2{,}114 &  526 &   0 \\
BUSI (breast ultrasound)         & Ultrasound / breast   &   518 &  129 &   0 \\
Br35H (brain tumor MRI)          & MRI / brain           &   800 &  100 &   0 \\
Head-CT hemorrhage               & CT / brain            &     0 &    0 & 200 \\
TN3K (thyroid US)                & Ultrasound / thyroid  &   800 &  200 &   0 \\
\bottomrule
\end{tabular}
\end{table}

\section{Additional ablations}
\label{app:extra-ablations}

This appendix collects the ablations that were summarized in
Section~\ref{sec:exp} but whose detailed numbers and figures are
deferred here for space.

\subsection{Capability supervision: granularity and sampling}
\label{app:extra-granularity}
\begin{figure}[ht]
  \centering
  \begin{minipage}[t]{0.5\linewidth}
    \centering
    \includegraphics[width=\linewidth]{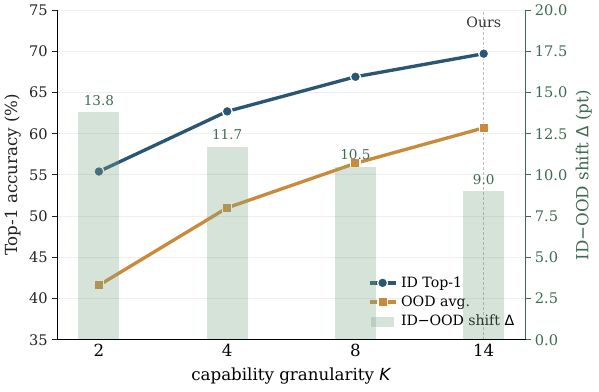}
  \end{minipage}
  \hfill
  \begin{minipage}[t]{0.46\linewidth}
    \centering
    \includegraphics[width=\linewidth]{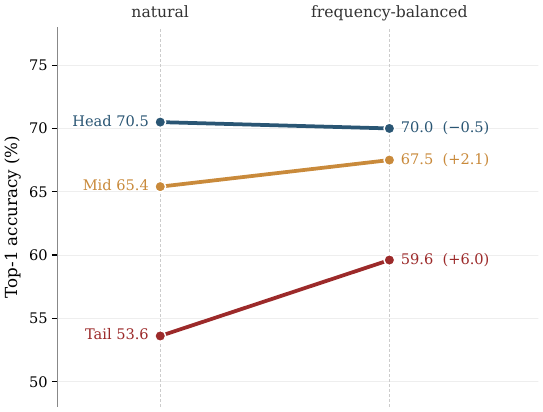}
  \end{minipage}
  \caption{\small Capability supervision ablations.
  \textbf{Left:} capability-granularity sweep. Lines show ID and OOD
  top-$1$ as a function of $K$; bars (right axis) show the
  $\mathrm{ID}-\mathrm{OOD}$ shift. Accuracy rises and the shift
  shrinks monotonically with $K$, with $K = 14$ as our operating
  point.
  \textbf{Right:} effect of frequency-balanced Stage-1 sampling. The
  slope chart connects per-tercile top-$1$ under natural sampling to
  the frequency-balanced sampler used by Med-Agentic; tail
  capabilities swing up sharply, while the head barely moves.}
  \label{fig:app-supervision}
\end{figure}

\paragraph{Capability granularity $K$.}
We vary $K \in \{2, 4, 8, 14\}$, starting from the two coarsest axes
(colour and density) and successively adding finer observations. As $K$
grows, both ID and OOD accuracy rise monotonically, and the
$\mathrm{ID}-\mathrm{OOD}$ shift narrows from $13.8$ to $9.0$ points
(Figure~\ref{fig:app-supervision}, left). Finer granularity gives the
student a richer basis for composition and shrinks the gap to
out-of-distribution diseases, supporting the design choice $K = 14$
adopted in Section~\ref{sec:taxonomy}.

\paragraph{Frequency-balanced Stage-1 sampling.}
\label{app:extra-sampling}
The per-capability training pool is long-tailed: the head five
capabilities exceed $13$k samples each, whereas density and radiopacity
have only $4.9$--$6.7$k. Figure~\ref{fig:app-supervision} (right)
compares natural sampling against the frequency-balanced sampler used
in Stage~1. Balancing lifts tail-capability top-$1$ by $+6.0$ points
while moving the head by only $-0.5$ points (within the seed-level
confidence interval), for a net gain of $+3.7$ points on ID and $+6.1$
points on OOD across the full test set.

\subsection{Capability-tagged CoT is a supervisable interface}
\label{app:extra-cot}
Table~\ref{tab:app-cot} ablates the CoT interface itself. Replacing
free-text CoT with capability-tagged CoT (without a segment-level loss)
improves top-$1$ by $+3.1$ points and grounding IoU by $0.037$, because
the tags render the trajectory parseable. Adding the segment-level loss
used by Ours further improves top-$1$ by $+2.4$ points and lifts IoU to
$0.431$. The $96.8\%$ well-formed rate confirms that the tag grammar is
a stable, learnable interface rather than a decorative artifact.

\begin{table}[ht]
\centering
\caption{\small CoT interface ablation. The segment-level loss is the
additional per-segment supervision introduced in
Section~\ref{sec:cot}.}
\label{tab:app-cot}
\small
\begin{tabular}{@{}llccc@{}}
\toprule
    & Variant                                     & Tag WF        & ID top-$1$    & IoU \\
\midrule
I0  & Free-text CoT (no tags)                      & \textsf{n/a}  & 64.2          & 0.358 \\
I1  & Tagged CoT, no segment-level loss            & 91.4          & 67.3          & 0.395 \\
I2  & Tagged CoT $+$ segment-level loss (Ours)     & \textbf{96.8} & \textbf{69.7} & \textbf{0.431} \\
\bottomrule
\end{tabular}
\end{table}

\subsection{Number of specialists $N$}
\label{app:extra-experts}
\begin{figure}[ht]
\centering
\includegraphics[width=0.55\linewidth]{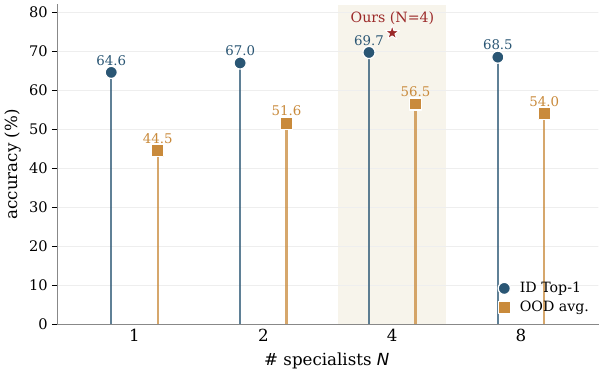}
\caption{\small Specialist-count ablation. Lollipop markers show ID top-$1$
(teal) and OOD average (ochre) at each $N$. $N = 4$ is the peak;
$N = 8$ hurts because per-specialist data thins out.}
\label{fig:app-experts}
\end{figure}

Figure~\ref{fig:app-experts} varies the number of Stage-3 specialists
$N \in \{1, 2, 4, 8\}$. At $N = 1$ the distillation degenerates to a
single generalist teacher and reproduces Stage-2 performance. Accuracy
rises through $N = 2$ and peaks at $N = 4$ (our choice: radiology,
fundus, RGB, ultrasound). Pushing to $N = 8$, which splits radiology
further into X-ray, CT and MRI specialists, costs $-2.5$ points on OOD
because per-specialist training data thins out. $N = 4$ is the right
balance for our corpus.

\subsection{Consequences of internalization}
\label{app:extra-internalization}
Table~\ref{tab:app-internalization} contrasts Ours against the
external-tool orchestration baseline on two system-level metrics. For
each sample, let $\mathcal{K}^{\star}(x) \subseteq \{\kappa_1, \ldots,
\kappa_{14}\}$ denote the set of capabilities that the host datasets
annotate as supported for image $x$, derived from the per-dataset
coverage table in Appendix~\ref{app:datasets}. The
out-of-$\mathcal{K}^{\star}$ call rate of the external-tool baseline is
$18.4\%$: the orchestrator frequently invokes a specialist that the
dataset says the image cannot support, because the tool API has no
direct gradient linking it to the diagnostic loss. Ours drives the
same rate to $4.0\%$ (a $4.6\times$ reduction) because capability tags
are generated by the student itself and receive segment-level
supervision. Ours is also end-to-end differentiable, which is a
necessary precondition for joint optimization of capability selection
and diagnostic accuracy.

\begin{table}[ht]
\centering
\caption{\small Consequences of internalization.
Out-of-$\mathcal{K}^{\star}$ denotes the fraction of capability
invocations whose capability is not annotated as supported for the
sample.}
\label{tab:app-internalization}
\small
\begin{tabular}{@{}lcc@{}}
\toprule
Metric                                            & External tool & \textbf{Med-Agentic (Ours)} \\
\midrule
Avg.\ \# capability calls per sample              & 4.5                  & \textbf{4.0} \\
Out-of-$\mathcal{K}^{\star}$ call rate            & 18.4\%               & \textbf{4.0\%} \\
End-to-end differentiable                          & \ding{55}            & \checkmark \\
\bottomrule
\end{tabular}
\end{table}

\subsection{Capability usage on OOD}
\label{app:extra-ood-usage}
\begin{figure}[ht]
\centering
\includegraphics[width=\linewidth]{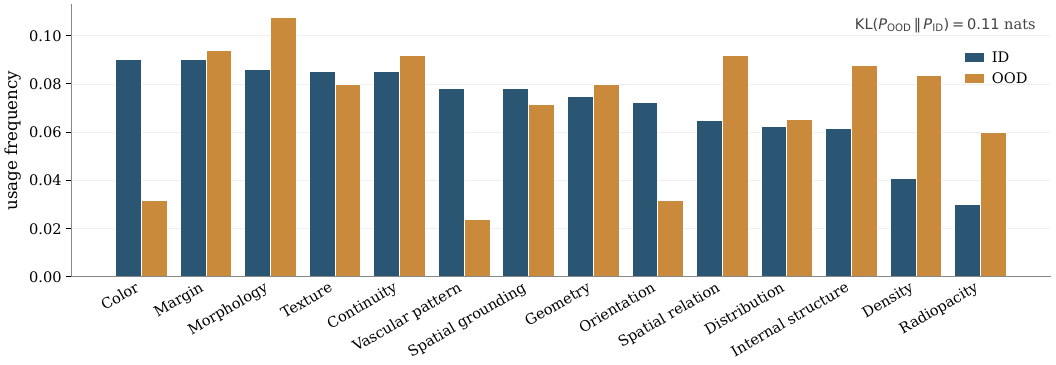}
\caption{\small Capability usage distribution on ID versus OOD. The two
distributions differ by $\mathrm{KL}(P_{\mathrm{OOD}} \,\|\,
P_{\mathrm{ID}}) \approx 0.11$~nats: the student reuses its capability
vocabulary on OOD rather than inventing a new one.}
\label{fig:app-ood-usage}
\end{figure}

Figure~\ref{fig:app-ood-usage} compares the per-capability usage
distribution on ID against the corresponding distribution on OOD. The
two distributions overlap heavily, with
$\mathrm{KL}(P_{\mathrm{OOD}} \,\|\, P_{\mathrm{ID}}) \approx 0.11$~nats:
on OOD the student reuses the same capability vocabulary it learned on
ID, only reweighting it in a physically coherent direction. CT-heavy
slices raise density, radiopacity, and internal-structure usage, while
RGB slices raise colour and vascular pattern, with the latter shift
diluted in the pooled average. This provides a mechanistic account of
how compositional generalization is realized in our model.

\section{Implementation details}
\label{app:impl}

This appendix collects the engineering details of the Med-OPD training
pipeline that were summarised in Section~\ref{sec:medopd}.

\subsection{Vision-encoder consistency}
The student and all four specialists share the same vision encoder
family, ensuring that the logit spaces compared in
Equation~\ref{eq:opd} are directly comparable.

\subsection{Teacher offload}
Because Equation~\ref{eq:opd} is evaluated on student-sampled
prefixes, we keep teachers frozen and cache only their last-layer
hidden states; logits are recomputed on-the-fly through a shared
prediction head, following recent OPD implementations~\citep{DBLP:journals/corr/abs-2306-13649}. This
keeps four specialists resident at modest memory overhead.

\subsection{Probe set for $\mathbf{W}^{\mathrm{cap}}$}
We construct a held-out probe set of approximately $50$ samples per
capability, disjoint from training, and estimate
$w^{\mathrm{cap}}_i[c]$ as the specialist's segment-level
GPT-$4$-as-judge score on capability $c$, normalized across $i$. The
matrix is computed once and frozen.

\subsection{Annotation reliability audit}
\label{app:annotation}
Capability annotations are partially derived from native dataset
labels and partially curated; this section reports inter-annotator
agreement on a stratified audit subset. We sample $100$ images,
stratified so that every anatomy and every capability is represented,
and have each image re-annotated by two independent
radiology-trained annotators, blinded to the dataset-native labels
and to each other. For each $(\text{image}, \text{capability})$ pair
we compare the automatically derived annotation (from native dataset
labels) against the two human annotations, and report three metrics.

\begin{itemize}[leftmargin=1.4em,itemsep=2pt,topsep=2pt]
\item \emph{Human--human agreement.} Cohen's $\kappa$ between the two
annotators on the binary "capability-observed" label, macro-averaged
over the $14$ capabilities. We observe $\bar\kappa = 0.81$, which is in
the ``substantial'' range under common interpretations.
\item \emph{Automatic--human agreement.} Cohen's $\kappa$ between the
dataset-label-derived capability annotation and the consensus of the
two humans. We observe $\bar\kappa = 0.78$, with the lowest
per-capability value ($\kappa = 0.66$) occurring on
\emph{texture\,/\,pattern}, where native dataset labels are least
specific.
\item \emph{Segment-boundary agreement.} On trajectories synthesized by
the Stage-$2$ annotator (Appendix~\ref{app:prompt6}--\ref{app:prompt7}),
we additionally measure the IoU between capability segments recovered
by the structural parser and those manually marked by the two
annotators. Mean segment IoU is $0.87$, confirming that the tag grammar
is identifiable from the raw text without ambiguity.
\end{itemize}

These numbers should be interpreted as an upper-bound on annotation
noise, since the audit set is stratified to cover both the head and
tail of the capability distribution.

\section{Training resources and hyperparameters}
\label{app:hparams}

This appendix lists the compute setup, optimizer settings, and
schedule used in each of the three training stages. All numbers
correspond to the run that produced the main-text results; the same
hyperparameters are used by every trained baseline (disease-level
SFT, fixed pipeline, external tool, plain GRPO) wherever the
corresponding stage applies, so that gaps in
Tables~\ref{tab:exp-id}--\ref{tab:exp-ood} reflect the training recipe
rather than tuning effort.

\subsection{Compute environment}
All training runs are executed on a single node with $8\times$
NVIDIA H200 GPUs ($141$\,GB HBM3e per device, NVLink-connected),
$2\times$ AMD EPYC CPUs, and $2$\,TB of system memory. The codebase
is built on PyTorch~$2.4$ with FSDP~(\textsc{full\_shard} +
\textsc{cpu\_offload} for teachers in Stage~$3$), FlashAttention-$2$,
and bf16 mixed precision throughout. The vision encoder is kept frozen
in Stages~$1$--$2$ and unfrozen with a $0.1\times$ learning-rate
multiplier in Stage~$3$.

\subsection{Backbone and shared settings}
The shared backbone for all trained methods is a Qwen3.5-9B
multimodal LLM; the tokenizer, vocabulary ($|\mathcal{V}|=152$k),
and vision encoder are shared across the student and the four
specialists, which is required for the full-vocabulary reverse-KL
objective (Equation~\ref{eq:opd}) to be well-defined.
Image inputs are resized so that the longer side is $896$ pixels and
patchified at stride $14$; the maximum text context length is set to
$8{,}192$ tokens, of which the structured-CoT trajectory occupies at
most $1{,}024$. All trained methods see the same $29{,}992$-sample
training pool (Appendix~\ref{app:datasets}) and the same
$\langle\textsc{think}\rangle\langle\textsc{answer}\rangle$ trajectory
template (Equation~\ref{eq:traj}), unless the ablation explicitly
removes the tag grammar (row~I0 of Table~\ref{tab:app-cot}).

\subsection{Per-stage hyperparameters}
Table~\ref{tab:app-hparams} reports the per-stage optimizer settings,
batch sizes, and schedules. Stage~$1$ and Stage~$2$
are standard supervised finetuning passes on the cold-start
trajectories described in Appendix~\ref{app:stage12}; Stage~$3$ is the
on-policy distillation procedure of Section~\ref{sec:medopd}.
``Specialist-forge'' is the lightweight per-cluster pass that produces
each $\pi_{T_i}$ before Med-OPD distillation begins
(Section~\ref{sec:medopd}); it consists of a short SFT pass on the
cluster's slice of the training pool followed by a brief GRPO step
with diagnostic accuracy as the scalar reward, and acts only as a
calibration of the four specialists rather than as the post-training
step itself.

\begin{table}[ht]
\centering
\small
\caption{\small Training hyperparameters across the three Med-Agentic stages.
``Effective batch'' is the global batch size after gradient
accumulation; ``Steps'' is the number of optimizer updates.
$\eta$ is the peak learning rate; the schedule is linear warmup over
the first $3\%$ of steps followed by cosine decay to $0.1\eta$ unless
stated otherwise. \emph{Specialist-forge} is the lightweight
per-cluster pass producing $\{\pi_{T_i}\}_{i=1}^{4}$, summed across
the four clusters.}
\label{tab:app-hparams}
\setlength{\tabcolsep}{4pt}
\begin{tabular}{@{}lccccc@{}}
\toprule
                                & Stage 1                & Stage 2                & Specialist-forge & Stage 3 (Med-OPD) \\
                                & Capability SFT         & Structured-CoT SFT     & SFT $+$ GRPO     & On-policy distill.\\
\midrule
Optimizer                       & AdamW                  & AdamW                  & AdamW     & AdamW \\
$\eta$ (peak)                   & $1{\times}10^{-5}$     & $5{\times}10^{-6}$     & $5{\times}10^{-6}$ & $2{\times}10^{-6}$ \\
$(\beta_1,\beta_2)$             & $(0.9, 0.95)$          & $(0.9, 0.95)$          & $(0.9, 0.95)$    & $(0.9, 0.95)$ \\
Weight decay                    & $0.05$                 & $0.05$                 & $0.05$           & $0.0$ \\
Grad clip                       & $1.0$                  & $1.0$                  & $1.0$            & $1.0$ \\
Schedule                        & cosine                 & cosine                 & cosine           & constant \\
Warmup                          & $3\%$                  & $3\%$                  & $3\%$            & $200$ steps \\
Effective batch (samples)       & $256$                  & $128$                  & $128$            & $64$ \\
Per-GPU micro-batch             & $4$                    & $2$                    & $2$              & $1$ \\
Grad-accum steps                & $8$                    & $8$                    & $8$              & $8$ \\
Sequence length (tokens)        & $2{,}048$              & $4{,}096$              & $4{,}096$        & $4{,}096$ \\
Steps                           & $9{,}000$              & $6{,}000$              & $4\times1{,}500$ & $5{,}000$ \\
Epochs (effective)              & 3            & 2            & 2 each & 1 \\
Vision encoder                  & frozen                 & frozen                 & frozen           & lr~$\times 0.1$ \\
KL temperatures                 & ---                    & ---                    & ---              & $\tau_1{=}1.0,\ \tau_2{=}0.5$ \\
GRPO group size $G$             & ---                    & ---                    & $8$              & ---  \\
GRPO KL coeff.\ $\beta$         & ---                    & ---                    & $0.04$           & ---  \\
Sampling temperature            & ---                    & ---                    & $1.0$            & $0.9$ \\
Top-$p$ (rollout)               & ---                    & ---                    & $0.95$           & $0.95$ \\
\bottomrule
\end{tabular}
\end{table}

Relative cost among stages is dominated by Med-OPD because each
student step requires a forward pass through all four frozen
specialists; capability SFT and structured-CoT SFT are significantly
cheaper, and specialist forging lies in between. Disease-level
multi-task SFT and the fixed-pipeline baseline use only Stage~$1$-style
supervised training on their respective targets; the external-tool
baseline reuses the four specialists from \emph{Specialist-forge} as
black-box callees and incurs no further training cost beyond a small
router-prompt fit. Plain GRPO replaces Stage~$3$ with a $5{,}000$-step
GRPO pass starting from $M_{\mathrm{cot}}$ using identical sampling
hyperparameters and a binary diagnostic-accuracy reward.

\subsection{Med-OPD-specific settings}
\label{app:medopd-settings}
Beyond the optimizer settings in Table~\ref{tab:app-hparams},
Med-OPD has three method-specific hyperparameters that are worth
isolating.
\begin{itemize}[leftmargin=1.4em,itemsep=2pt,topsep=2pt]
\item \emph{Routing temperatures.} We set $\tau_1=1.0$ in
  Equation~\ref{eq:wimg} and $\tau_2=0.5$ in Equation~\ref{eq:wcap}.
  Sweeping $\tau_1\in\{0.5,1.0,2.0\}$ and $\tau_2\in\{0.25,0.5,1.0\}$
  on a $1{,}000$-sample validation slice changes ID top-$1$ by at
  most $\pm 0.4$ pt and OOD by at most $\pm 0.6$ pt; we therefore
  adopt the geometric centre and do not retune per ablation.
\item \emph{Centroids $\mu_i$.} The vision-encoder centroid of each
  specialist's training shard is computed once at the start of
  Stage~$3$ over $2{,}048$ randomly sampled cluster images, and is
  frozen thereafter.
\item \emph{Capability competence matrix
  $\mathbf{W}^{\mathrm{cap}}\!\in\!\mathbb{R}_{+}^{4\times 15}$.}
  Estimated once on the held-out probe set
  (Appendix~\ref{app:impl}); raw entries lie in $[0.31, 0.92]$
  with row-wise dynamic range $\approx 2.4\times$, which is what
  drives the per-token reweighting in Equation~\ref{eq:dualfactor}.
\end{itemize}
All other Stage~$3$ choices, including the full-vocabulary reverse-KL
objective and the stop-gradient treatment of routing weights, are
specified in Section~\ref{sec:medopd} and Appendix~\ref{app:proofs}.

\subsection{Reproducibility checklist}
Every trained number reported in the main text is the mean over three
seeds ($\{17, 42, 2026\}$); standard deviations are within $\pm 0.5$
ID top-$1$ and $\pm 0.8$ OOD top-$1$ across all trained controls.
Evaluation uses greedy decoding with the same trajectory template as
training and a single deterministic forward pass; no test-time tool
calls or self-consistency aggregation are performed.

\section{Synthetic data construction}
\label{app:syndata}

This appendix details the construction of the multimodal,
capability-tagged training corpus. The seven verbatim prompts used to
elicit each annotation type are listed in Appendix~\ref{app:prompts};
here we report the data flow, scale, filtering, and quality control
that turn those prompts into the $29{,}992$-sample training pool of
Section~\ref{sec:exp-setup}.

\subsection{Annotator stack}
All synthetic annotations are produced by
\textbf{Gemini-3-Flash}\footnote{The same model family is used for
both the per-capability annotations of Appendices~\ref{app:prompt1}--\ref{app:prompt5}
and the two-node agentic CoT annotator of
Appendices~\ref{app:prompt6}--\ref{app:prompt7}; for brevity, the
prompt listings refer to it as ``the annotator''.} accessed through
its multimodal endpoint at temperature $0.2$, top-$p$ $0.95$, and a
$1{,}024$-token output cap (the $\approx 50$-token soft cap inside
the prompts is enforced as a JSON-schema post-condition rather than
by truncation). Each request carries the patient image at original
resolution and a structured JSON instruction; replies that fail JSON
parsing or violate the anti-leakage constraints
(Appendix~\ref{app:prompts}) are rejected and re-queried up to two
times. The per-capability competence matrix
$\mathbf{W}^{\mathrm{cap}}$ used by Med-OPD
(Section~\ref{sec:medopd}) is in turn estimated by a separate, frozen
GPT-$4$ judge on the held-out probe set; this evaluator role is
disjoint from the Gemini-3-Flash annotator role and is not used to
generate any training token.

\subsection{Pipeline}
For each of the $112{,}821$ candidate images aggregated from the
$30$ source datasets (after deduplication on a perceptual hash), we
construct training instances as follows:
\begin{enumerate}[leftmargin=1.4em,itemsep=2pt,topsep=2pt]
\item \textbf{Spatial grounding (per image).} Exactly one of
  Prompts~$1$--$3$ in Appendix~\ref{app:prompts} is dispatched
  depending on whether the source dataset already provides a
  ground-truth lesion box (Prompt~$2$), provides a confirmed-negative
  label (Prompt~$3$), or provides neither (Prompt~$1$); the response
  yields a $50$-token blind analysis and a normalized
  $[\![0,1000]\!]^{4}$ box.
\item \textbf{Fine-grained attribute capabilities (per image, per
  capability).} For each of the $13$ non-grounding capabilities
  whose source dataset is annotated in
  \texttt{attribute\_stats.txt} as supporting capability $\kappa_i$
  on this image, we dispatch Prompt~$5$ (positive samples, with the
  ROI from Step~$1$) or Prompt~$4$ (negatives). The same image
  therefore generates a variable number of single-capability
  $(\textit{analysis},\textit{label})$ pairs, on average
  $5.4$ across the corpus.
\item \textbf{Two-node agentic CoT (per image).} Prompts~$6$ and~$7$
  are dispatched in sequence: the planning node selects the subset of
  capability tools that a blind examiner would exercise, the
  diagnostic node consumes the executed tool log and emits the final
  $\langle\textsc{think}\rangle\langle\textsc{answer}\rangle$
  trajectory of the form in Equation~\ref{eq:traj}. The
  anti-leakage constraint is enforced jointly by the system prompts
  in Appendices~\ref{app:prompt6}--\ref{app:prompt7}.
\end{enumerate}
The output is a per-image bundle containing one grounding annotation,
$1$--$13$ attribute annotations, and one structured-CoT trajectory.
Stage~$1$ training data are sourced from the per-capability bundles
(Step~2) plus a copy of the grounding bundle (Step~1); Stage~$2$
training data are sourced from the trajectory bundles (Step~3).

\subsection{Scale and yield}
Table~\ref{tab:app-syndata-scale} summarizes the size of each step in
the pipeline. The aggregate Gemini-3-Flash budget is approximately
$1.04$\,M successful multimodal calls, with a global yield of
$87.4\%$ after JSON-validity filtering, anti-leakage filtering, and
deduplication.

\begin{table}[ht]
\centering
\small
\caption{\small Scale of the synthetic-data pipeline. Per-step yields are
fractions of \emph{successful, retained} responses out of dispatched
calls (including the two automatic retries described in
Section~\ref{app:syndata}). The grounding line counts the union of
Prompts~$1$--$3$. ``Trajectories'' are joint planning$+$diagnostic
outputs: a row is retained only if both nodes pass.}
\label{tab:app-syndata-scale}
\setlength{\tabcolsep}{5pt}
\begin{tabular}{@{}lrrrr@{}}
\toprule
Pipeline step                                    & Calls dispatched & Calls retained & Yield      & Output unit \\
\midrule
P1--P3 \emph{Spatial grounding}                  & $112{,}821$       & $103{,}946$    & $92.1\%$   & one analysis $+$ box per image \\
P4--P5 \emph{Per-capability attributes}          & $704{,}589$       & $612{,}801$    & $87.0\%$   & $1$--$13$ pairs per image \\
P6--P7 \emph{Two-node CoT trajectories}          & $225{,}642$       & $179{,}480$    & $79.5\%$   & one trajectory per image \\
\midrule
\textbf{Total Gemini-3-Flash calls}              & $\mathbf{1{,}043{,}052}$ & $\mathbf{896{,}227}$ & $\mathbf{85.9\%}$ & --- \\
\bottomrule
\end{tabular}
\end{table}

\subsection{Filtering and deduplication}
Three filters run on every retained response:

\paragraph{Schema and length.}
We require valid JSON and enforce the $\approx 50$-token soft cap
on the \emph{analysis} field with a hard cap of $80$ tokens
(violations are re-queried up to two times, then dropped). The
\emph{label} field is constrained to $1$--$3$ tokens and rejected if
it intersects a curated $317$-term blacklist of disease names; this
guards against leakage from the clinical context provided in the
prompt. Roughly $7.6\%$ of first-pass responses are dropped at this
stage.

\paragraph{Anti-leakage regex.}
A second filter scans the \emph{analysis} field for $74$ phrases
that betray the hidden context (\texttt{within the box}, \texttt{the
specified area}, \texttt{as the label says}, \texttt{the target
diagnosis}, $\ldots$); offending samples are rejected and re-queried.
This filter rejects roughly $4.1\%$ of responses, concentrated in the
first weeks of pipeline development; the converged rejection rate on
the final corpus is $0.9\%$.

\paragraph{Trajectory consistency (CoT only).}
For Step~$3$, we additionally require that the diagnostic-node
trajectory references only capability tags that the planning node
previously selected, that all capability tags belong to the
$14$-axis taxonomy, and that the
$\langle\textsc{think}\rangle/\langle\textsc{answer}\rangle$ grammar
parses cleanly. Trajectories that fail any of the three checks are
dropped; the structural pass-rate is $89.1\%$.

\subsection{Human spot-check}
On top of the inter-annotator audit reported in
Appendix~\ref{app:annotation}, a $200$-image stratified sample is
re-annotated by two clinicians blinded to the synthetic labels and
to each other. Three quantities are computed.
\begin{itemize}[leftmargin=1.4em,itemsep=2pt,topsep=2pt]
\item \emph{Capability-label agreement.} Cohen's $\kappa$ between the
  Gemini-3-Flash label and the consensus of the two clinicians,
  macro-averaged across the $14$ capabilities, is $\bar\kappa = 0.78$
  (range $[0.66, 0.91]$); the lowest value is on
  \emph{texture\,/\,pattern}, where source-dataset labels are least
  specific.
\item \emph{Trajectory faithfulness.} On the $200$-trajectory subset,
  $186$ ($93.0\%$) are judged ``faithful and forward-looking'' (i.e.\
  the reasoning could plausibly precede the diagnosis without
  retrospective leakage), $11$ are judged borderline, and $3$
  exhibit hindsight phrasing despite passing the regex filter. We
  remove all such hindsight cases and their image-level neighbours
  ($168$ trajectories total) from the final training pool.
\item \emph{Bounding-box recall.} The clinician-redrawn lesion boxes
  achieve $\mathrm{IoU}\ge 0.5$ with the Gemini-3-Flash boxes on
  $87.5\%$ of positive samples, with median IoU $0.71$.
\end{itemize}
These numbers are consistent with the inter-annotator estimates of
Appendix~\ref{app:annotation} and place an empirical upper bound on
the noise contributed by the synthetic-data pipeline; they should be
read as the floor of the supervision signal that the student model
inherits.

\section{Stage~1 and Stage~2 training details}
\label{app:stage12}

This appendix expands on the two cold-start stages of
Section~\ref{sec:pipeline}, namely \emph{capability SFT} and
\emph{structured-CoT SFT}. We give the precise objectives, the data
mixtures, the segment-level loss referenced in
Section~\ref{sec:cot}, and the design choices that distinguish each
stage from a generic instruction-tuning pass.

\subsection{Stage 1: capability SFT}
\label{app:stage1}

\paragraph{Data.}
Stage~$1$ uses the per-capability bundles from
Appendix~\ref{app:syndata}, formatted as instruction-following triples
$(x, q_i, y_i)$, where $x$ is the image, $q_i$ is the capability-key
prompt
\texttt{"Use \{$\kappa_i$\} to describe this image."}
or its grounding-specific variant
\texttt{"Localize the region that supports any abnormal observation."}
for the spatial-grounding capability, and $y_i$ is the
single-capability target produced by Prompts~$1$--$5$ of
Appendix~\ref{app:prompts}. After filtering, the Stage-1 pool contains
$612{,}801$ attribute pairs over the $13$ attribute capabilities and
$103{,}946$ grounding pairs, a total of $716{,}747$ training
instances.

\paragraph{Frequency-balanced sampling.}
Per-capability counts are long-tailed (Figure~\ref{fig:capfreq}); we
counteract this with frequency-inverse sampling. Concretely, at each
step we sample a capability index $i\sim p(i)\propto n_i^{-\alpha}$
with $\alpha=0.5$, where $n_i$ is the number of Stage-$1$ pairs for
capability $\kappa_i$, and then sample a $(x,q_i,y_i)$ triple
uniformly from that capability's pool. The exponent $\alpha=0.5$
recovers the head$\to$tail tradeoff visualized in
Figure~\ref{fig:app-supervision} (right) and is selected by validation
sweep over $\alpha\in\{0.0,0.25,0.5,0.75,1.0\}$ on the Stage-$1$
holdout slice.

\paragraph{Objective.}
Let $y_i = (y_{i,1},\ldots,y_{i,L})$ be the target token sequence.
Stage-$1$ minimizes the standard token-level cross-entropy on the
target sequence only,
\begin{equation}
\label{eq:stage1-loss}
\mathcal{L}_{\mathrm{S1}}(\theta)
= \mathbb{E}_{(x,q_i,y_i)}\!\left[
-\sum_{\ell=1}^{L} \log \pi_\theta(y_{i,\ell}\mid y_{i,<\ell}, x, q_i)
\right],
\end{equation}
with the prompt tokens masked out of the loss and a uniform weight on
each target token. We do not introduce any auxiliary objective at this
stage: the goal is fluency on each capability in isolation, not their
composition. The output of Stage~$1$ is the checkpoint
$M_{\mathrm{base}}$ used to initialize Stage~$2$.

\paragraph{Curriculum within Stage~1.}
We split Stage~$1$ into two phases of $4{,}500$ optimizer steps each.
Phase~$1\mathrm{a}$ trains exclusively on attribute capabilities
(Prompts~$4$--$5$); Phase~$1\mathrm{b}$ adds the grounding capability
(Prompts~$1$--$3$) and continues training. This curriculum was
adopted after we observed that mixing the spatial-grounding sequences,
which contain numeric box tokens, with the prose-style attribute
sequences from the start slowed convergence on attribute fluency by
roughly $20\%$ in early steps; once attributes stabilize the joint
mixture trains stably.

\subsection{Stage 2: structured-CoT SFT}
\label{app:stage2}

\paragraph{Data.}
Stage~$2$ uses the trajectory bundles from
Appendix~\ref{app:syndata}, retained after the structural pass-rate
and human spot-check filters in Section~\ref{app:syndata}. The Stage-2
pool contains $179{,}480$ trajectories of the form in
Equation~\ref{eq:traj}, with on average $5.4$ capability segments
per trajectory and a tail at $11$. Each trajectory is paired with the
diagnostic question $q$ and the original image $x$.

\paragraph{Two complementary trajectory sources.}
As stated in Section~\ref{sec:pipeline}, trajectories are constructed
in two ways and both contribute to the Stage-$2$ pool.
\begin{itemize}[leftmargin=1.4em,itemsep=2pt,topsep=2pt]
\item \emph{Annotator-generated} ($75.0\%$ of trajectories). The
two-node agentic annotator (Prompts~$6$--$7$,
Appendices~\ref{app:prompt6}--\ref{app:prompt7}) emits the trajectory
end-to-end conditional on the image, the dataset's native labels,
and the executed per-capability tool log; the anti-leakage
constraints in those prompts force the trajectory to read as
forward-looking.
\item \emph{Programmatic} ($25.0\%$ of trajectories). For samples
where the source dataset's per-capability annotations are
sufficiently complete, we order the segments along a clinician-authored
template (e.g.\ \emph{global $\to$ local}, \emph{detection $\to$
attribute $\to$ grounding}) and concatenate the per-capability
analyses produced in Stage~$1$. This branch contributes no new
content but enforces a clinically plausible ordering on which the
model's composition policy is warm-started.
\end{itemize}
The $3{:}1$ mixture is preserved within every minibatch via
weighted sampling.

\paragraph{Token-level cross-entropy.}
Let $\tau = (z_1,\ldots,z_T)$ be a Stage-$2$ trajectory, of the form
\[
\tau = \big[\langle\textsc{think}\rangle\;\kappa_{i_1}{:}o_1,\ldots,\kappa_{i_m}{:}o_m\;\langle/\textsc{think}\rangle\;\langle\textsc{answer}\rangle\;d\;\langle/\textsc{answer}\rangle\big].
\]
The base trajectory loss is the standard masked cross-entropy on
$\tau$ given the image $x$ and question $q$,
\begin{equation}
\label{eq:stage2-base}
\mathcal{L}_{\mathrm{S2}}^{\mathrm{tok}}(\theta)
= \mathbb{E}_{(x,q,\tau)}\!\left[
-\sum_{t=1}^{T}\mathbb{1}\{z_t \in \mathrm{response}\}\;
\log \pi_\theta(z_t \mid z_{<t}, x, q)
\right],
\end{equation}
with the question and any system tokens masked out. Without any
further regularization, this objective already produces well-formed
tagged CoT $91.4\%$ of the time (row~I1 in Table~\ref{tab:app-cot}).

\paragraph{Segment-level loss.}
Let $\mathrm{seg}(\tau) = \{S_1,\ldots,S_m\}$ be the partition of
$\tau$ induced by the structural parser, where each segment $S_j$
spans the tokens of one $\kappa_{i_j}{:}o_j$ block (or the single
\textsc{answer} block). For each segment we define a per-segment loss
\begin{equation*}
\ell_j(\theta)
= -\frac{1}{|S_j|}\sum_{t\in S_j}
\log \pi_\theta(z_t\mid z_{<t}, x, q),
\end{equation*}
i.e.\ the average negative log-likelihood within that segment. The
\emph{segment-level loss} is the unweighted mean over segments,
\begin{equation}
\label{eq:stage2-seg}
\mathcal{L}_{\mathrm{S2}}^{\mathrm{seg}}(\theta)
= \mathbb{E}_{(x,q,\tau)}\!\left[\frac{1}{m}\sum_{j=1}^{m} \ell_j(\theta)\right].
\end{equation}
Because long segments contribute the same weight as short ones,
Equation~\ref{eq:stage2-seg} prevents the few capability segments
with longer associated observations (typically morphology and
internal-structure) from dominating the gradient, and forces the
student to fit short, rare segments such as radiopacity equally well.
The full Stage-$2$ objective is the convex combination
\begin{equation}
\label{eq:stage2-full}
\mathcal{L}_{\mathrm{S2}}(\theta)
= (1-\lambda)\,\mathcal{L}_{\mathrm{S2}}^{\mathrm{tok}}(\theta)
+ \lambda\,\mathcal{L}_{\mathrm{S2}}^{\mathrm{seg}}(\theta),
\qquad \lambda = 0.3,
\end{equation}
selected by validation sweep over $\lambda\in\{0,0.1,0.3,0.5,0.7\}$.
Setting $\lambda=0$ recovers row~I1 of Table~\ref{tab:app-cot}; the
default $\lambda=0.3$ corresponds to row~I2 (Ours) and improves
top-$1$ by $+2.4$ pt, IoU by $+0.036$, and CoT well-formed rate by
$+5.4$ pt.

\paragraph{Identifiability of segments.}
The structural parser implementing $\mathrm{seg}(\tau)$ is a single
regular expression keyed on the $14$ capability-tag tokens and the
\textsc{think}/\textsc{answer} delimiters. On a $500$-trajectory
audit set we observed zero parse failures and a mean segment-IoU of
$0.87$ against manual segmentation, consistent with the figure
reported in Appendix~\ref{app:annotation}; we therefore treat
$\mathrm{seg}(\tau)$ as identifiable and use it without uncertainty
weighting in Equation~\ref{eq:stage2-seg}.

\paragraph{Output of Stage~2.}
The output checkpoint $M_{\mathrm{cot}}$ achieves a $94.2\%$
well-formed CoT rate on the Stage-$2$ holdout slice and $64.6$ ID
top-$1$ (row~S2 of Table~\ref{tab:exp-id} pipeline dissection in the
main text); it is the cold-start initializer for both the four
specialists $\{\pi_{T_i}\}_{i=1}^{4}$ and the student $\pi_\theta$
of Stage~$3$.

\subsection{What is \emph{not} done in Stages~1--2}
For clarity, two design choices that are deliberately deferred to
Stage~$3$:
\begin{enumerate}[leftmargin=1.4em,itemsep=2pt,topsep=2pt]
\item Stages~$1$--$2$ do not optimize any reward derived from the
final diagnosis $d$. The label $d$ appears in the trajectory only as
a supervised target, never as a scalar reward; reward-weighted
updates are reserved for the per-cluster GRPO calibration of
specialists and, as a contrast, for the plain-GRPO baseline.
\item Stages~$1$--$2$ do not introduce any teacher distinct from the
student. The four specialists $\{\pi_{T_i}\}$ and the dual-factor
routing of Equation~\ref{eq:dualfactor} only become active in
Stage~$3$. Consequently, $M_{\mathrm{cot}}$ is the strongest
single-checkpoint variant of our model that does not require multiple
forward passes per training token.
\end{enumerate}

\section{Proofs and additional theoretical remarks}
\label{app:proofs}

We use the following notation. For a token prefix $z_{<t}$ and input
$x$, write
\begin{align*}
H_i(t,x,z_{<t};\theta)
&= \nabla_\theta
D_{\mathrm{KL}}\!\left(
\pi_\theta(\cdot\mid z_{<t},x)\,\|\,\pi_{T_i}(\cdot\mid z_{<t},x)
\right),\\
K_i(t,x,z_{<t};\theta)
&= D_{\mathrm{KL}}\!\left(
\pi_\theta(\cdot\mid z_{<t},x)\,\|\,\pi_{T_i}(\cdot\mid z_{<t},x)
\right).
\end{align*}
The teachers are frozen, and the routing coefficients are treated as
mixture weights, equivalently stop-gradient coefficients, when taking
the student update. This is the estimator implemented in
Equation~\ref{eq:opd}; if one deliberately differentiated through the
routing network, extra $\nabla_\theta w_i$ terms would define a different
objective. We assume the KL gradients are integrable and the routing
weights lie in the simplex, both of which hold under the softmax
parameterization in Equations~\ref{eq:wimg}-\ref{eq:dualfactor}.

\paragraph{Proposition~\ref{prop:generalize} (strict generalization).}
Let $\mathcal{W}_{\mathrm{prompt}}$ be the per-prompt routing family,
\begin{align*}
\mathcal{W}_{\mathrm{prompt}}
= \Big\{\, w: w_i(t\mid x,\tau)=\bar w_i(x),\
\bar w(x)\in\Delta^{N-1}\,\Big\},
\end{align*}
and let $\mathcal{W}_{\mathrm{dual}}$ be the family induced by
Equations~\ref{eq:wimg}-\ref{eq:dualfactor},
\begin{align*}
w_i(t\mid x,\tau)
= \frac{w_i^{\mathrm{img}}(x)\,w_i^{\mathrm{cap}}[c(z_t)]}
       {\sum_{j=1}^{N} w_j^{\mathrm{img}}(x)\,w_j^{\mathrm{cap}}[c(z_t)]}.
\end{align*}
The objective in Equation~\ref{eq:opd} is completely determined by the
choice of $w\in\mathcal{W}_{\mathrm{dual}}$:
\begin{align*}
\mathcal{L}_{w}(\theta)
= \mathbb{E}_{(x,q)}\mathbb{E}_{\tau\sim\pi_\theta}
\left[\sum_{t=1}^{T}\sum_{i=1}^{N}w_i(t\mid x,\tau)
K_i(t,x,z_{<t};\theta)\right].
\end{align*}

First, $\mathcal{W}_{\mathrm{prompt}}\subseteq\mathcal{W}_{\mathrm{dual}}$.
Take any per-prompt weights $\bar w(x)\in\Delta^{N-1}$ and set
$w_i^{\mathrm{img}}(x)\equiv\bar w_i(x)$. Set the capability competence
matrix to be column-constant, e.g.\
$\mathbf{W}^{\mathrm{cap}}_{i,c}\equiv 1$ for all $i,c$. Then
Equation~\ref{eq:wcap} gives $w_i^{\mathrm{cap}}[c]\equiv 1/N$, and
Equation~\ref{eq:dualfactor} becomes
\begin{align*}
w_i(t\mid x,\tau)
&= \frac{\bar w_i(x)\,(1/N)}{\sum_{j=1}^{N}\bar w_j(x)\,(1/N)}
 = \bar w_i(x),
\end{align*}
for every token $t$ and every trajectory $\tau$. Substituting this into
$\mathcal{L}_w$ recovers exactly the per-prompt OPD objective
$\mathcal{L}_{\mathrm{prompt}}$.

Second, the containment is strict whenever the specialists are
non-degenerate. It suffices to construct a routing rule in
$\mathcal{W}_{\mathrm{dual}}$ that no per-prompt rule can express. Let
$N=2$ and fix one image $x$ whose trajectory can contain two capability
tags $a$ and $b$ with positive probability. Choose
$w^{\mathrm{img}}(x)=(1/2,1/2)$ and choose the capability factor so that
\begin{align*}
w^{\mathrm{cap}}[a]=(3/4,1/4),
\qquad
w^{\mathrm{cap}}[b]=(1/4,3/4).
\end{align*}
Such factors are feasible because a softmax over positive competence
scores can realize any interior point of the simplex by setting the
log-scores proportionally to the desired logits. The resulting
dual-factor weights satisfy
\begin{align*}
w_1(t\mid x,c(z_t)=a)=3/4,
\qquad
w_1(t\mid x,c(z_t)=b)=1/4.
\end{align*}
Any per-prompt routing rule must assign a single value
$\bar w_1(x)$ to all tokens of the same input $x$, hence would require
simultaneously $\bar w_1(x)=3/4$ and $\bar w_1(x)=1/4$, a contradiction.
Therefore $\mathcal{W}_{\mathrm{dual}}\nsubseteq
\mathcal{W}_{\mathrm{prompt}}$. If the two teachers induce different
KL values on at least one of the two token classes, the corresponding
objectives also differ, so the inclusion of objective families is
strict rather than merely notational. Thus dual-factor Med-OPD
strictly generalizes per-prompt OPD.\hfill$\square$

\paragraph{Proposition~\ref{prop:unbiased} (unbiasedness).}
The statement compares dual-factor routing to the matched-marginal
per-prompt estimator. The short statement in Section~\ref{sec:medopd}
writes $\bar w_i(x)$ for readability; the precise token-level quantity
conditions on the sampled prefix,
\begin{align*}
\bar w_i(x,z_{<t})
= \mathbb{E}_{c_t\sim p(\cdot\mid x,z_{<t})}
\left[w_i(t\mid x,\tau)\right],
\qquad c_t=c(z_t).
\end{align*}
Averaging $\bar w_i(x,z_{<t})$ over prefixes gives the prompt-level
marginal $\bar w_i(x)$ used in the main text. Define the corresponding
matched per-prompt gradient at the same prefix as
\begin{align*}
\bar g_t(\theta)
= \sum_{i=1}^{N}\bar w_i(x,z_{<t})\,H_i(t,x,z_{<t};\theta),
\end{align*}
and the dual-factor gradient estimator as
\begin{align*}
g_t(\theta)
= \sum_{i=1}^{N}w_i(t\mid x,\tau)\,H_i(t,x,z_{<t};\theta).
\end{align*}
Condition on $(x,z_{<t})$. Once the prefix is fixed,
$H_i(t,x,z_{<t};\theta)$ is fixed with respect to the current capability
random variable $c_t$; only the routing weight changes with $c_t$. Hence
linearity of conditional expectation gives
\begin{align*}
\mathbb{E}_{c_t\mid x,z_{<t}}\!\left[g_t(\theta)\right]
&= \mathbb{E}_{c_t\mid x,z_{<t}}\!\left[
   \sum_{i=1}^{N}w_i(t\mid x,\tau)\,H_i(t,x,z_{<t};\theta)
   \right]\\
&= \sum_{i=1}^{N}
   \mathbb{E}_{c_t\mid x,z_{<t}}\!\left[w_i(t\mid x,\tau)\right]
   H_i(t,x,z_{<t};\theta)\\
&= \sum_{i=1}^{N}\bar w_i(x,z_{<t})\,
   H_i(t,x,z_{<t};\theta)\\
&= \bar g_t(\theta).
\end{align*}
Taking the outer expectation over $(x,q)$, the sampled trajectory prefix
$z_{<t}\sim\pi_\theta$, and the token index $t$ yields
\begin{align*}
\mathbb{E}_{x,q,z_{<t},c_t}\!\left[g_t(\theta)\right]
= \mathbb{E}_{x,q,z_{<t}}\!\left[\bar g_t(\theta)\right].
\end{align*}
Equivalently, after marginalizing the prefix distribution, the concise
form in Equation~\ref{eq:unbiased} holds:
\begin{align*}
\mathbb{E}[g_t(\theta)\mid x]
= \mathbb{E}[\bar g_t(\theta)\mid x].
\end{align*}
Thus token-level dual-factor routing changes the conditional variance
and expressivity of the teacher mixture, but not the expected teacher-KL
gradient relative to a per-prompt comparator with matched marginals.
The required regularity conditions are mild: bounded simplex weights,
finite KL gradients, full support over realized capability tags, frozen
teacher policies, and stop-gradient routing coefficients. The softmax
parameterization gives bounded positive weights and full support, so
these conditions are satisfied by Med-OPD.\hfill$\square$

\section{Annotation prompts}
\label{app:prompts}

For reproducibility, we list verbatim the seven annotation prompts used to
construct the Med-Agentic training corpus. Prompts~1-3 target the
spatial-grounding (detection) capability under three sample regimes
(unlabeled, positive with ground-truth box, negative); prompts~4-5 target
the remaining $13$ fine-grained visual meta-capabilities for negative and
positive samples respectively; prompts~6-7 implement the two-node agentic
CoT annotator (planning node and diagnostic node) used to synthesize the
capability-tagged trajectories distilled in Stage~3. All prompts share a
common anti-leakage design: annotation targets must read as blind,
forward-looking observations that do not reference the underlying labels,
bounding boxes, or target diagnoses that condition the generator.

\subsection{Spatial grounding: detection without ground-truth box}
\label{app:prompt1}

Used to jointly produce a blind visual analysis and a tight bounding box
for samples that lack a detection annotation.

\begin{lstlisting}[style=promptstyle]
messages=[{"role": "system", "content": (
        "You are a top-tier medical image annotation expert. Locate and bound the target lesion region based on the instructions and clinical context. "
        "[Core Principle]: Single compact bounding box (Minimum Bounding Box). Tightly fit lesion edges. "
        "[Anti-Leakage Constraint]: The 'analysis' field will be used as a training target. It MUST read as an independent, blind radiological observation. STRICTLY PROHIBIT any references to the provided disease labels, clinical context, or the task instructions. Just describe the visual findings directly.Describe the lesion's appearance and location relative to anatomical landmarks to justify the coordinates. "
        "[STRICT LENGTH LIMIT]: The 'analysis' field MUST be extremely concise, bounded around 50 tokens (approx. 30-40 words). Long paragraphs are strictly prohibited! "
        "[Output Format]: Strict JSON (no Markdown): {\"analysis\": \"[Around 50 tokens] Blind visual analysis and localization logic\", \"label\": \"[xmin, ymin, xmax, ymax]\"} "
        "[Coordinates]: 1000x1000 normalized space (origin top-left)."
    )},
    {"role": "user", "content": [
        {"type": "text", "text": f"{config[item['source_dataset']]['prompts']['detection']}\n\n[Clinical Context]\n" + "\n---\n".join([f"Disease Category: {disease_label}\nOverview: {overview}\nAs shown in the image, the current sample is diagnosed with: {name}. Typical visual characteristics of this state are: {desc}"])},
        {"type": "image_url", "image_url": {"url": f"data:image/png;base64,{img_b64}"}}
    ]}]
\end{lstlisting}

\subsection{Spatial grounding: positive sample with known box}
\label{app:prompt2}

Used when a ground-truth lesion box is already available; the annotator
produces only the reasoning chain that justifies the given region.

\begin{lstlisting}[style=promptstyle]
messages=[{"role": "system", "content": (
    "You are a top-tier medical image annotation expert. The target lesion bounding box is ALREADY KNOWN and provided in the prompt. Your task is ONLY to provide the reasoning chain (analysis) that leads to this bounding box. "
    "[Anti-Leakage Constraint]: The 'analysis' field will be used as a training target. It MUST read as an independent, blind radiological observation. STRICTLY PROHIBIT any references to the provided bounding box coordinates, disease labels, or clinical context. Act as if you discovered the lesion yourself. Do NOT say 'Within the box', 'The specified area', or 'Based on the label'. Just describe the visual findings directly. "
    "[STRICT LENGTH LIMIT]: The 'analysis' field MUST be extremely concise, bounded around 50 tokens (approx. 30-40 words). Long paragraphs are strictly prohibited! "
    "[Output Format]: Strict JSON (no Markdown): {\"analysis\": \"[Around 50 tokens] Blind visual analysis logic explaining the lesion\", \"label\": \"KNOWN\"}"
)},
    {"role": "user", "content": [
        {"type": "text", "text": f"{base_det_prompt}\n\n[Known Target Lesion Bounding Box]: {gt_box}\n\n[Clinical Context]\n" + "\n---\n".join([f"Disease Category: {disease_label}\nOverview: {overview}\nAs shown in the image, the current sample is diagnosed with: {name}. Typical visual characteristics of this state are: {desc}"])},
        {"type": "image_url", "image_url": {"url": f"data:image/png;base64,{img_b64}"}}
    ]}]
\end{lstlisting}

\subsection{Spatial grounding: negative sample}
\label{app:prompt3}

Used for confirmed negative (normal) samples; the annotator explains why
the anatomy is healthy without referencing the given label.

\begin{lstlisting}[style=promptstyle]
messages=[{"role": "system", "content": (
    "You are a top-tier medical image annotation expert. It is ALREADY KNOWN that this sample is completely normal (negative) with no abnormal lesion areas. Your task is ONLY to provide the reasoning chain (analysis) confirming why it is normal.Focus on describing the integrity and healthy appearance of the specific anatomical structures mentioned in the instructions. "
    "[Anti-Leakage Constraint]: The 'analysis' field will be used as a training target. It MUST read as an independent, blind radiological observation. STRICTLY PROHIBIT any references to the provided disease labels or clinical context. Do NOT say 'As the label says normal'. Just describe the visual findings of healthy anatomy directly. "
    "[STRICT LENGTH LIMIT]: The 'analysis' field MUST be extremely concise, bounded around 50 tokens (approx. 30-40 words). Long paragraphs are strictly prohibited! "
    "[Output Format]: Strict JSON (no Markdown): {\"analysis\": \"[Around 50 tokens] Blind visual analysis confirming healthy anatomy\", \"label\": \"NORMAL\"}"
)},
    {"role": "user", "content": [
        {"type": "text", "text": f"{config[item['source_dataset']]['prompts']['detection']}\n\n[Clinical Context]\n" + "\n---\n".join([f"Disease Category: {disease_label}\nOverview: {overview}\nAs shown in the image, the current sample is diagnosed with: {name}. Typical visual characteristics of this state are: {desc}"])},
        {"type": "image_url", "image_url": {"url": f"data:image/png;base64,{img_b64}"}}
    ]}]
\end{lstlisting}

\subsection{Fine-grained attribute capabilities: negative sample}
\label{app:prompt4}

Used to annotate one of the $13$ non-grounding capabilities on a normal
sample, strictly along the requested analytical dimension.

\begin{lstlisting}[style=promptstyle]
messages=[{"role": "system", "content": (
    "You are a meticulous medical visual feature extraction expert. Evaluate the image strictly along the SPECIFIC analytical dimension requested in the user prompt (e.g., color, geometry, margins, continuity). "
    "[Task Alignment]: Do NOT describe the general 'most prominent' feature of the lesion. You MUST answer the specific question asked. If asked about color, your label MUST be a color descriptor. If asked about geometry, it MUST be a geometric descriptor.The 'analysis' must provide evidence ONLY for this single dimension. "
    "[Single Tag Constraint]: The 'label' field must strictly match the semantic grain and brevity of the <label reference examples>. It should be a single, dimension-aligned atomic phrase of similar length to the examples (typically 1-3 words)."
    "[Normal/Negative Cases]: If the sample is healthy with no lesions, output a label that describes normal anatomy strictly WITHIN the requested dimension. Do not output generic phrases like 'Normal background'. "
    "[Core Constraint]: Describe ONLY physical features. NO disease names. "
    "[Anti-Leakage Constraint]: The 'analysis' field will be used as a training target. It MUST read as a blind, objective observation. STRICTLY PROHIBIT meta-terms that reveal the prompt's context, such as 'The ROI', 'within the bounding box', 'specified coordinates', or 'clinical label'. Use direct anatomical subjects (e.g., 'A tissue mass is visible...', NOT 'Within the ROI, a mass is visible...'). "
    "[STRICT LENGTH LIMIT]: The 'analysis' field MUST be extremely concise, bounded around 50 tokens (approx. 30-40 words). Long paragraphs are strictly prohibited! "
    "[Output Format]: Strict JSON (no Markdown): {\"analysis\": \"[Around 50 tokens] Blind visual analysis focusing ONLY on the requested dimension\", \"label\": \"[ONLY ONE] dimension-aligned feature phrase\"}"
)},
    {"role": "user", "content": [
        {"type": "text", "text": f"{config[item['source_dataset']]['prompts']['attributes'][attr]}\n\n[Clinical Context]\n" + "\n---\n".join([f"Disease Category: {disease_label}\nOverview: {overview}\nAs shown in the image, the current sample is diagnosed with: {name}. Typical visual characteristics of this state are: {desc}"]) + "\n\n[Visual Reference Guide]: This sample is NORMAL with NO lesions. Please evaluate the general healthy anatomy or the entire image context."},
        {"type": "image_url", "image_url": {"url": f"data:image/png;base64,{img_b64}"}}
    ]}]
\end{lstlisting}

\subsection{Fine-grained attribute capabilities: positive sample}
\label{app:prompt5}

Same system instruction as Appendix~\ref{app:prompt4}; the user turn
additionally passes a focal ROI so that the annotator concentrates on the
lesion region.

\begin{lstlisting}[style=promptstyle]
messages=[{"role": "system", "content": (
    "You are a meticulous medical visual feature extraction expert. Evaluate the image strictly along the SPECIFIC analytical dimension requested in the user prompt (e.g., color, geometry, margins, continuity). "
    "[Task Alignment]: Do NOT describe the general 'most prominent' feature of the lesion. You MUST answer the specific question asked. If asked about color, your label MUST be a color descriptor. If asked about geometry, it MUST be a geometric descriptor.The 'analysis' must provide evidence ONLY for this single dimension. "
    "[Single Tag Constraint]: The 'label' field must strictly match the semantic grain and brevity of the <label reference examples>. It should be a single, dimension-aligned atomic phrase of similar length to the examples (typically 1-3 words)."
    "[Normal/Negative Cases]: If the sample is healthy with no lesions, output a label that describes normal anatomy strictly WITHIN the requested dimension. Do not output generic phrases like 'Normal background'. "
    "[Core Constraint]: Describe ONLY physical features. NO disease names. "
    "[Anti-Leakage Constraint]: The 'analysis' field will be used as a training target. It MUST read as a blind, objective observation. STRICTLY PROHIBIT meta-terms that reveal the prompt's context, such as 'The ROI', 'within the bounding box', 'specified coordinates', or 'clinical label'. Use direct anatomical subjects (e.g., 'A tissue mass is visible...', NOT 'Within the ROI, a mass is visible...'). "
    "[STRICT LENGTH LIMIT]: The 'analysis' field MUST be extremely concise, bounded around 50 tokens (approx. 30-40 words). Long paragraphs are strictly prohibited! "
    "[Output Format]: Strict JSON (no Markdown): {\"analysis\": \"[Around 50 tokens] Blind visual analysis focusing ONLY on the requested dimension\", \"label\": \"[ONLY ONE] dimension-aligned feature phrase\"}"
)},
    {"role": "user", "content": [
        {"type": "text", "text": f"{config[item['source_dataset']]['prompts']['attributes'][attr]}\n\n[Clinical Context]\n" + "\n---\n".join([f"Disease Category: {disease_label}\nOverview: {overview}\nAs shown in the image, the current sample is diagnosed with: {name}. Typical visual characteristics of this state are: {desc}"]) + f"\n\n[Visual Reference Guide]: 1. Space: 1000x1000 normalized (Origin top-left).\n2. ROI: {pred_box} [xmin, ymin, xmax, ymax]. Focus primarily on this bounding box."},
        {"type": "image_url", "image_url": {"url": f"data:image/png;base64,{img_b64}"}}
    ]}]
\end{lstlisting}

\subsection{Agentic CoT annotator: planning node}
\label{app:prompt6}

Selects the subset of capability tools that a blind examiner would
exercise, conditioned on the target diagnosis yet strictly forbidden from
revealing it.

\begin{lstlisting}[style=promptstyle]
messages=[{"role": "system", "content": (
    "You are the Planning Node of an AI Medical Agent. Your task is to select the best combination of analytical tools to investigate an unknown medical condition.\n"
    "[Available Tools]: You will be provided with a list of available visual analysis tools for this specific sample AND their actual execution results.\n"
    "[Target Diagnosis]: You will be provided with the final expected diagnosis to guide your tool selection secretly, BUT YOU MUST HIDE THIS KNOWLEDGE.\n"
    "[Anti-Leakage Constraint]: STRICTLY PROHIBIT mentioning the Target Diagnosis, disease names, or labels in your <plan>. You must act as if you are performing a completely blind, exploratory examination. Use general exploratory phrasing (e.g., 'To evaluate the retinal health and check for structural anomalies...', NEVER 'To confirm wet AMD...').\n"
    "[Strategy]:\n"
    "1. Do NOT blindly use all available tools. Select ONLY the tools necessary to investigate and deduce the condition, increasing the diversity of reasoning paths.\n"
    "2. 'Lesion_Detection' is NOT mandatory (e.g., massive global abnormalities might not need localization). BUT if you choose to use it, it MUST be the very first tool called.\n"
    "3. Call each tool at most once.\n"
    "4. Choose as many necessary tools as possible to build a comprehensive diagnostic case.\n"
    "[CRITICAL CONCEALMENT CONSTRAINT - READ CAREFULLY]:\n"
    "Although you are provided with the Target Diagnosis and the Tool Results, YOU MUST COMPLETELY HIDE THIS FACT in your output. Your <plan> must read as a forward-looking, exploratory hypothesis from a blind doctor.\n"
    "- NEVER use hindsight logic (e.g., 'Since the result shows fluid...' or 'To confirm the wet AMD...').\n"
    "- ALWAYS use exploratory logic (e.g., 'To investigate the potential presence of fluid...', 'To evaluate structural integrity...').\n"
    "[Output Format]: STRICT JSON format exactly as follows:\n"
    "{\n"
    "  \"diagnosis_plan\": \"<plan> Explain your exploratory strategy in 1-2 sentences. You MUST explicitly mention the tools you want to use exactly in this format: [Tool_Call: Exact_Tool_Name]. </plan>\"\n"
    "}"
)},
    {"role": "user", "content": f"[Target Diagnosis]: {disease_name}\n[Available Tools and Results]: {tools_with_results}\nPlease formulate the diagnosis plan."}]
\end{lstlisting}

\subsection{Agentic CoT annotator: diagnostic node}
\label{app:prompt7}

Synthesizes the executed tool log into a final reasoning chain that
concludes with the target diagnosis while appearing to read only the
textual tool reports.

\begin{lstlisting}[style=promptstyle]
messages=[{"role": "system", "content": (
    "You are the Diagnostic Node of an AI Medical Agent. Your task is to synthesize the executed tool results into a final conclusion.\n"
    "[Inputs]: You will be provided with the Agent's Plan, the Executed Tool Log, and the Target Diagnosis, and the original medical image.\n"
    "[Task Objective]: Write the final reasoning chain (<cot>) that logically synthesizes the <result>s to justify the Target Diagnosis.\n"
    "[CRITICAL VISUAL CONCEALMENT CONSTRAINT - READ CAREFULLY]:\n"
    "You are provided with the original medical image to guarantee the absolute accuracy of your diagnosis. HOWEVER, your written reasoning (<cot>) must be strictly reverse-engineered to appear as if you ONLY read the text in the [Executed Tool Log].\n"
    "- NEVER mention 'visible in the image', 'looking at the scan', or 'the image shows'.\n"
    "- NEVER describe any anatomical or visual details (e.g., colors, specific shapes, exact locations) that are NOT explicitly mentioned in the <result> or <cot> of the [Executed Tool Log].\n"
    "- Frame your reasoning entirely as a synthesis of the tool reports.\n"
    "[Constraints]: \n"
    "- Speak directly as the AI expert reviewing the findings. Do not refer to 'The log says'.\n"
    "- [Anti-Leakage Constraint]: Ensure the reasoning feels like a natural, forward-deduction process from visual evidence to diagnosis. Do NOT reveal that you were given the diagnosis in advance (e.g., NEVER say 'This confirms the given target diagnosis...'). Just state the findings and conclude naturally.\n"
    "- Keep the <cot> concise (around 40 words).\n"
    "[Output Format]: STRICT JSON format exactly as follows:\n"
    "{\n"
    "  \"diagnostic_cot\": \"<cot> [Your synthesized clinical reasoning leading to the conclusion here] </cot>\"\n"
    "}"
)},
    {"role": "user", "content": f"[Target Diagnosis]: {disease_name}\n[Plan]:\n{diagnosis_plan}\n\n[Executed Tool Log]:\n" + "\n\n".join([f"[Tool_Call: {tool}]\n<cot> {cot} </cot>\n<result> {res} </result>"]) + "\n\nPlease output the final synthesis."}]
\end{lstlisting}

\section{Targeted validation and transfer boundaries}
\label{app:validation}

This supplement reports additional evaluations from the author response. The original paper tables are retained above; the closed-set re-evaluation and image/annotation accounting below are reported separately so that different evaluation protocols and counting units are not combined. The following experiments separate four questions: whether diagnosis-conditioned supervision accounts for the gains, whether generated observations affect answer decoding, whether known capabilities can be recombined in unseen combinations, and where a fixed capability basis or internalized knowledge is insufficient. Descriptive capability-use patterns are reported in Section~\ref{sec:exp-interp}.

\subsection{Diagnosis-withheld supervision}
\label{app:blind}
We regenerate annotations over the same $29{,}992$ training-image IDs, keeping the annotation model/checkpoint, decoding settings, trajectory/token budget, initialization, and training schedule fixed. In the \emph{node-blind} condition, the target diagnosis is withheld from both planning and diagnostic nodes. The stricter \emph{fully diagnosis-withheld} condition also removes disease name, category, clinical overview, typical visual characteristics, and positive/normal status from every per-capability prompt. The ground-truth label is appended to the \texttt{<answer>} field only after the generated \texttt{<think>} trajectory is frozen; it is never used for selection, regeneration, or filtering. All conditions use the same $6{,}033$-image ID test set and $516$-image OOD accuracy benchmark.

\begin{table}[ht]
\centering\small
\caption{Diagnosis conditioning control. OOD is the unweighted macro-average top-$1$ across the three accuracy slices.}
\label{tab:blind}
\begin{tabular}{@{}lrr@{}}
\toprule
Supervision & ID top-$1$ (\%) & OOD top-$1$ (\%) \\
\midrule
Disease-level SFT reference & 60.1 & 35.8 \\
Original diagnosis-conditioned & 69.7 & 56.5 \\
Planning/diagnostic nodes blind & 68.4 & 54.6 \\
Fully diagnosis-withheld & 67.2 & 52.9 \\
\bottomrule
\end{tabular}
\end{table}

Fully withholding diagnosis reduces ID/OOD accuracy by $2.5/3.6$ points, but retains $74.0\%/82.6\%$ of the original gain over disease-level SFT. Its capability selection has mean Jaccard similarity $0.76$ to the conditioned counterpart. Two clinical annotators, blinded to target diagnosis and generation condition, judge $85.0\%$ of a modality-stratified sample of $200$ fully blind observations to be image-supported (Cohen's $\kappa=0.74$). Hindsight conditioning therefore contributes to performance without explaining all gains. This audit does not establish that every generated token is grounded or causally faithful.

\subsection{Controlled capability-segment interventions}
\label{app:interventions}
On the $6{,}033$ ID cases, we fix the image, query, checkpoint, decoding configuration, and all non-target CoT segments. We edit one target segment and regenerate only the final answer with greedy decoding. The irrelevant-deletion control is length-matched; contradictory replacements minimally edit an observation, with a clinically consistent alternative specified before evaluating the changed answer. Restoring the exact original segment and serialization provides a reversibility control.

\begin{table}[ht]
\centering\small
\caption{Paired segment interventions. Flip rate is relative to the intact answer; $\Delta$GT log-odds is the change in ground-truth log-odds. Direction is the fraction of changed predictions moving toward the prespecified alternative for contradictory replacement.}
\label{tab:interventions}
\setlength{\tabcolsep}{5pt}
\begin{tabular}{@{}lrrrr@{}}
\toprule
Intervention & Acc. (\%) & Flip (\%) & $\Delta$GT log-odds & Direction (\%) \\
\midrule
Intact & 69.7 & 0.0 & 0.00 & n/a \\
Tag-only perturbation & 68.9 & 3.2 & $-0.09$ & n/a \\
Irrelevant-segment deletion & 68.4 & 4.6 & $-0.14$ & n/a \\
Relevant-segment deletion & 61.8 & 13.4 & $-0.62$ & n/a \\
Contradictory replacement & 55.6 & 22.8 & $-1.08$ & 72.4 \\
Restore exact original & 69.7 & 0.0 & 0.00 & n/a \\
\bottomrule
\end{tabular}
\end{table}

Relevant deletion causes a larger accuracy drop than irrelevant deletion ($7.9$ vs.\ $1.3$ points). Contradictory replacement has the strongest effect, while tag-only changes have a comparatively small effect. These interventions support downstream answer dependence on observation content, rather than only tag strings. They do not establish independent latent modules, nor do they prove that the original observations were constructed without diagnosis-conditioned hindsight; the control in Appendix~\ref{app:blind} addresses that complementary issue.

\subsection{Taxonomy distinguishability and label-count controls}
\label{app:taxonomy-controls}
After removing capability names and diagnosis terms, grouped leave-one-dataset-out capability prediction achieves $68.4\%$ macro-$F_1$, versus $7.1\%$ under label permutation. It remains $62.7\%$ after controlling for anatomy and modality. The largest confusion is morphology versus margin ($16.4\%$), consistent with partial overlap. We additionally match samples, observation-token budget, and optimization steps across semantic, merged, and random pseudo-split taxonomies.

\begin{table}[ht]
\centering\small
\caption{Matched-budget taxonomy controls. Increasing label count alone does not improve this comparison; the results do not establish a unique atomic decomposition.}
\label{tab:taxonomy-controls}
\begin{tabular}{@{}lrr@{}}
\toprule
Taxonomy & ID top-$1$ (\%) & OOD macro-average (\%) \\
\midrule
Semantic-14 & 69.7 & 56.5 \\
Merged-11 & 67.6 & 52.4 \\
Random pseudo-split-28 & 69.3 & 55.9 \\
\bottomrule
\end{tabular}
\end{table}

Merging the disputed dimensions reduces ID/OOD accuracy by $2.1/4.1$ points, whereas meaningless pseudo-tags do not improve performance. Alongside the prediction and intervention tests, this supports empirically distinguishable, reusable observation dimensions at the chosen reporting granularity, without claiming statistical independence or neural orthogonality.

\subsection{Combination-disjoint evaluation}
\label{app:combinations}
We hold out $28$ capability pairs and $36$ triples spanning all modality groups, with every constituent capability individually observed during training. In the pair evaluation, a held-out pair never co-occurs in any multi-capability training stage. In the triple evaluation, the target triple is absent while all three constituent pairs remain observed. Seen controls are matched by capability count, anatomy, modality, and diagnosis. Pair and triple holdouts are distinct evaluation conditions; they should not be read as ordinary test slicing of an unrestricted training run.

The pair evaluation uses $1{,}200$ seen-control and $1{,}200$ novel-combination cases; the triple evaluation uses $900$ of each. All four subsets are mutually disjoint subsets of the $6{,}033$-image ID test set, add no images, and are never optimized on. Methods use the same condition-specific split, training budget, and three seeds. Table~\ref{tab:combinations} reports mean $\pm$ standard deviation.

\begin{table}[ht]
\centering\small
\caption{Seen versus novel capability combinations (top-$1$, \%). Seen-to-novel drops for Med-Agentic are $4.3/6.6$ points on pairs/triples, versus $5.4/8.3$ for GRPO.}
\label{tab:combinations}
\setlength{\tabcolsep}{6pt}
\begin{tabular}{@{}lcccc@{}}
\toprule
Method & Seen pair & Novel pair & Seen triple & Novel triple \\
\midrule
Disease-level SFT & $60.0\pm0.5$ & $52.3\pm0.7$ & $59.3\pm0.6$ & $47.9\pm0.8$ \\
Fixed pipeline & $62.2\pm0.6$ & $55.5\pm0.8$ & $61.4\pm0.5$ & $51.0\pm0.7$ \\
Matched tools & $63.3\pm0.5$ & $57.4\pm0.7$ & $62.6\pm0.6$ & $53.6\pm0.8$ \\
Plain GRPO & $65.7\pm0.4$ & $60.3\pm0.6$ & $64.8\pm0.5$ & $56.5\pm0.7$ \\
Med-Agentic & $69.4\pm0.4$ & $65.1\pm0.5$ & $69.0\pm0.4$ & $62.4\pm0.6$ \\
\bottomrule
\end{tabular}
\end{table}

Med-Agentic exceeds GRPO by $4.8$ points on novel pairs and $5.9$ on novel triples. This is direct evidence for recombining known capabilities within the fixed vocabulary. It does not demonstrate capability discovery or unrestricted transfer to new observation dimensions.

\subsection{Taxonomy-independent PCam evaluation}
\label{app:pcam}
We evaluate the full class-balanced $32{,}768$-image official PatchCamelyon (PCam)~\citep{veeling2018pcam} test split for lymph-node metastasis histopathology. Neither this modality nor its metastatic/benign lymph-node disease family and schema was used to construct the original taxonomy. PCam is excluded from training, model selection, and prompt tuning. Before inspecting outputs, we freeze all $14$ capability definitions, parameters, prompts, preprocessing, and the binary candidate labels (metastatic tissue versus benign tissue). No capability is added, renamed, merged, or revised using PCam.

\begin{table}[ht]
\centering\small
\caption{Taxonomy-independent PCam top-$1$. The balanced binary benchmark has a $50\%$ chance reference; the relative gain should not obscure the low absolute accuracy.}
\label{tab:pcam}
\begin{tabular}{@{}lr@{}}
\toprule
Method & Top-$1$ (\%) \\
\midrule
Disease-level SFT & 32.7 \\
Fixed pipeline & 36.2 \\
Matched tools & 40.7 \\
Plain GRPO & 46.1 \\
Med-Agentic & 52.1 \\
\bottomrule
\end{tabular}
\end{table}

Two pathology-trained clinicians, blinded to model outputs, independently assess a prespecified balanced $2{,}000$-image subset for expressibility using only the frozen basis. Coverage is $89.3\%$ ($\kappa=0.79$). On that audited subset, Med-Agentic accuracy is $55.4\%$ for fully expressible cases and $25.0\%$ for cases needing observations outside the basis. The relative gain over GRPO is $6.0$ points, but $52.1\%$ overall accuracy is only slightly above chance. This is boundary evidence: the frozen taxonomy provides limited transfer but is insufficient for reliable histopathology diagnosis. A new modality can still require adaptation, and missing observational dimensions require expert supervision and retraining.

\subsection{Guideline retrieval and complementary external knowledge}
\label{app:rag}
The original matched-tools control invokes our own specialists and isolates orchestration under matched task knowledge. To test genuinely additional knowledge, Guideline-RAG uses the same controller backbone with public WHO, ACR, and ESGE guidance snapshots dated May 31, 2026, excluded from task-specific training and synthetic annotation. Documents are divided into $512$-token chunks with $64$-token overlap, indexed by hybrid BM25 and BGE-M3~\citep{chen2025m3} retrieval, and combined by reciprocal-rank fusion. The controller forms its query from image-derived capability observations and the candidate-label list before answering; neither the ground-truth label nor an initial prediction enters the query. The top five passages are given to the frozen controller.

Within the OOD accuracy benchmark, two clinicians independently select $150$ knowledge-dependent cases before inspecting outputs. Inclusion requires an adjudicated guidance passage containing a diagnostic distinction absent from the task-specific sources. The shuffled-evidence control replaces passages within the same source and length stratum, preserving top-$k$ and context length. Absence from task-specific sources does not establish absence from backbone pretraining.

\begin{table}[ht]
\centering\small
\caption{Retrieval comparison. OOD is the three-slice macro-average; KD is accuracy on the $150$ knowledge-dependent cases. Latencies are median and interquartile range.}
\label{tab:rag}
\begin{tabular}{@{}lrrrl@{}}
\toprule
Method & OOD (\%) & KD (\%) & Median (s) & IQR (s) \\
\midrule
Med-Agentic & 56.5 & 47.3 & 1.2 & 1.0--1.5 \\
Guideline-RAG & 54.1 & 58.7 & 6.7 & 5.8--7.9 \\
RAG + shuffled evidence & 49.8 & 49.3 & 6.6 & 5.7--7.8 \\
Med-Agentic + RAG & 60.5 & 63.3 & 7.1 & 6.2--8.4 \\
\bottomrule
\end{tabular}
\end{table}

Guideline-RAG exceeds Med-Agentic by $11.4$ points on knowledge-dependent cases (paired-bootstrap $95\%$ CI: $+4.1$ to $+18.7$), and the advantage largely disappears under shuffled evidence. The hybrid is strongest overall, but has approximately $5.9\times$ the median latency of Med-Agentic alone. Thus efficient internalized reasoning under stable task knowledge and external access to additional guidance are complementary capabilities. Selective use of retrieved evidence is also emphasized by SURf~\citep{DBLP:conf/emnlp/SunZZSQ024}, which improves multimodal retrieval utilization and robustness to irrelevant or misleading references.

\section{Standardized zero-shot evaluation and data accounting}
\label{app:eval-protocol}

\subsection{Closed-set evaluation protocol}
Every zero-shot model receives the same dataset-specific candidate labels; label order is randomized by a fixed per-example seed shared across models. There are no demonstrations or task-specific hints. The prompt requests one final label in machine-readable form. Normalization, fixed before evaluation, handles Unicode, case, surrounding whitespace, and punctuation, then exact-matches a predeclared alias table. Multiple, invalid, or unmatched answers count as incorrect, with no manual or model-based semantic correction. Decoding uses temperature $0$, top-$p=1.0$, maximum $512$ output tokens, and one attempt per image. API models were accessed on July 31, 2026. Zero-shot results are single-run evaluations; three-seed averages apply to the trained-method comparisons.

\begin{table}[ht]
\centering\small
\caption{Re-evaluated zero-shot contextual references under the standardized closed-set protocol. The checkpoint/endpoint strings identify the evaluated systems. Only top-$1$ and OOD macro-average were reported for this re-evaluation; earlier-protocol slice or auxiliary scores are not mixed into these results.}
\label{tab:zero-protocol}
\setlength{\tabcolsep}{4pt}
\begin{tabular}{@{}p{0.23\linewidth}p{0.49\linewidth}rr@{}}
\toprule
Model & Endpoint/checkpoint & ID (\%) & OOD (\%) \\
\midrule
Gemini 2.5 Flash & \nolinkurl{gemini-2.5-flash} & 39.2 & 26.0 \\
Gemini 3 Flash & \nolinkurl{gemini-3-flash-preview} & 43.8 & 28.5 \\
GPT-5.5 & \nolinkurl{gpt-5.5-2026-04-23} & 46.0 & 30.0 \\
Claude Sonnet 4.6 & \nolinkurl{claude-sonnet-4-6} & 44.5 & 29.2 \\
Qwen3.5-9B & \nolinkurl{Qwen/Qwen3.5-9B} & 31.4 & 18.6 \\
Qwen3.5-27B & \nolinkurl{Qwen/Qwen3.5-27B} & 38.6 & 24.5 \\
Hulu-Med-7B & \nolinkurl{ZJU-AI4H/Hulu-Med-7B} & 34.0 & 20.7 \\
HuatuoGPT-Vision-7B & \nolinkurl{FreedomIntelligence/HuatuoGPT-Vision-7B-hf} & 36.5 & 22.4 \\
\bottomrule
\end{tabular}
\end{table}

Generic zero-shot label-only outputs are not scored against our tagged-CoT grammar or grounding format. The primary matched-training comparison remains $69.7/56.5$ for Med-Agentic versus $65.8/50.3$ for plain GRPO, rather than a headline against differently supervised zero-shot systems.

\subsection{Image and annotation ledger}
\label{app:data-ledger}
The $119{,}373$-image original corpus comprises $29{,}992$ training, $82{,}829$ reserve, $6{,}033$ ID test, and $519$ OOD-analysis images. Annotation candidates are training plus reserve ($112{,}821$); neither ID nor OOD images are optimization inputs. The three-slice OOD accuracy benchmark is a $516$-image subset of the analysis pool. The remaining three images are single-image label-schema holdouts from ISIC-MILK10k, PAD-UFES-20, and Dental Panoramic Radiography, used only for coverage/complexity analysis.

\begin{table}[ht]
\centering\small
\caption{Stage-wise accounting. Candidate-wide retained counts include reserve annotations; consumed counts refer only to selected training-image IDs. Node responses and complete trajectories are different units.}
\label{tab:data-ledger}
\begin{tabular}{@{}p{0.21\linewidth}p{0.20\linewidth}p{0.27\linewidth}p{0.23\linewidth}@{}}
\toprule
Stage & Image scope & Retained across candidates & Used for optimization \\
\midrule
Candidate generation & 112,821 unique images & Training + reserve candidates & 29,992 image IDs \\
Stage 1 & Same candidate pool & 716,747 image--capability instances & 180,752 instances on training IDs \\
Stage 2 & Same candidate pool & 179,480 node responses; 89,740 paired trajectories & 29,992 trajectories on training IDs \\
Stage 3 & 29,992 training images & On-policy student trajectories & Training IDs only \\
\bottomrule
\end{tabular}
\end{table}

The $4{,}200$ pair/triple test cases are mutually disjoint subsets of the ID set, not new images. The PCam evaluation adds the full $32{,}768$-image official test split from a $35$th dataset; its $2{,}000$-image audit is a subset. The resulting total across all analyses is $152{,}141$ unique images from $35$ sources. PCam contributes no taxonomy-design, training, selection, or prompt-tuning data.

\end{document}